\documentclass[10pt,twocolumn,letterpaper]{article}

\usepackage{iccv}
\usepackage{times}
\usepackage{epsfig}
\usepackage{graphicx}
\usepackage{amsmath}
\usepackage{amssymb}
\usepackage{algorithm,algpseudocode}
\usepackage{tabularx}
\usepackage{booktabs}
\usepackage{multirow} 

\usepackage{enumitem}

\usepackage[pagebackref=true,breaklinks=true,letterpaper=true,colorlinks,bookmarks=false]{hyperref}

\iccvfinalcopy 

\def\iccvPaperID{2978} 

\ificcvfinal\fi

\begin{document}
\definecolor{veg}{RGB}{107, 142, 35}
\title{Re-distributing Biased Pseudo Labels for Semi-supervised Semantic Segmentation: A Baseline Investigation}


\author{
Ruifei He$^{1,2}$\thanks{equal contribution}, \  Jihan Yang$^{1*}$, \ Xiaojuan Qi$^{1}$\thanks{corresponding author}\\
$^{1}$The University of Hong Kong \ $^{2}$Zhejiang University \\
{\tt\small rfhe@zju.edu.cn, \{jhyang, xjqi\}@eee.hku.hk}
}
\maketitle
\ificcvfinal\thispagestyle{empty}\fi

\begin{abstract}
    \vspace{-0.2cm}
   While self-training has advanced semi-supervised semantic segmentation, it severely suffers from the long-tailed class distribution on real-world semantic segmentation datasets that make the pseudo-labeled data bias toward majority classes.
   In this paper, we present a simple and yet effective Distribution Alignment and Random Sampling \textbf{(DARS)} method to produce unbiased pseudo labels that match the true class distribution estimated from the labeled data.
   Besides, we also contribute a progressive data augmentation and labeling strategy to facilitate model training with pseudo-labeled data.
   Experiments on both Cityscapes and PASCAL VOC 2012 datasets demonstrate the effectiveness of our approach. Albeit simple, our method performs favorably in comparison with state-of-the-art approaches. Code will be available at \url{https://github.com/CVMI-Lab/DARS}.
\end{abstract}
 
\vspace{-0.5cm}
\section{Introduction}
\vspace{-0.1cm}
Recent years have witnessed the great success of deep convolutional neural network (DCNNs) in semantic segmentation \cite{sermanet2013overfeat,long2015fully,chen2016attention,liu2015parsenet,zhao2017pyramid}.
The success, however, heavily relies on a large number of training data with accurate pixel-level human annotations, which are prohibitively expensive and time-consuming to collect.

Semi-supervised learning (SSL) provides a promising path \cite{xie2019unsupervised,tarvainen2017mean,laine2016temporal,xie2020self,sohn2020fixmatch}  to ease human annotation burden by using a small amount of labeled data in conjunction with a large amount of unlabeled data to obtain an accurate model.
In this regard, self-training, alternating between generating pseudo labels for unlabeled data using model predictions and training the model with pseudo-labeled data, is a classic and effective approach for semi-supervised learning and has obtained state-of-the-art results \cite{feng2020semi,chen2020semi,zoph2020rethinking}  in semi-supervised semantic segmentation with DCNNs.

\begin{figure}
\begin{center}
\includegraphics[width=1.0\linewidth]{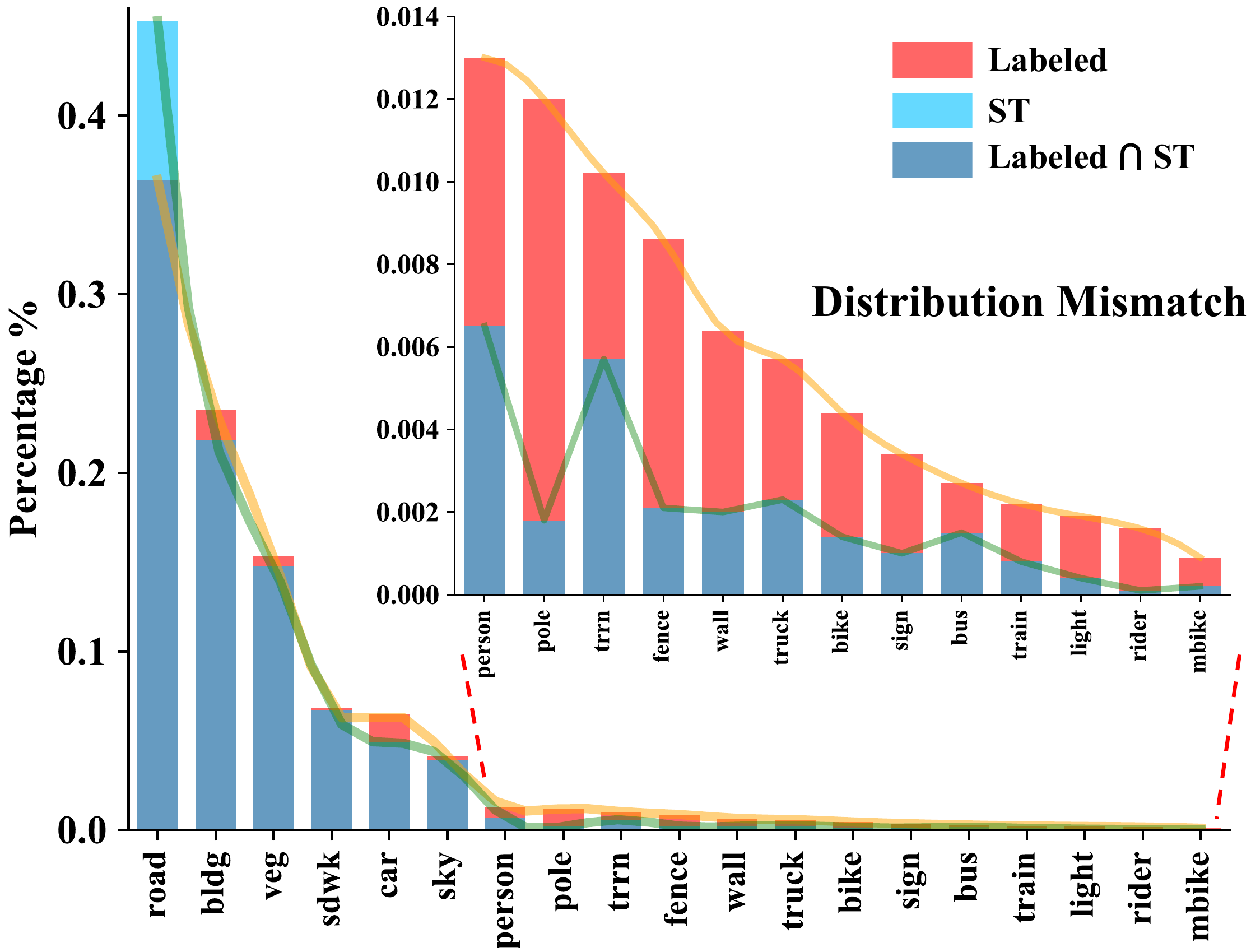} 
\end{center}
\vspace{-0.4cm}
   \caption{Class distribution mismatch on the Cityscapes dataset~\cite{cordts2016cityscapes}. 'Labeled' and 'ST' denote the class distribution of true labels in the labeled set and pseudo labels produced by ST. We line up percentages of each class for better visualization. }
\label{fig:distribution}
\vspace{-0.6cm}
\end{figure}

\vspace{-0.1cm}
\vspace{0.1in}\noindent\textbf{Motivations.} Despite the encouraging results, most of the previous self-training approaches \cite{zoph2020rethinking,zhu2020improving,lian2019constructing,zhang2020transferring,sohn2020fixmatch} assume a class-balanced data distribution and hence adopt a single confidence thresholding \textbf{(ST)} scheme to produce the pseudo labels ({\ie} pixels with prediction confidence score exceeding a pre-defined threshold are pseudo-labeled) to guarantee pseudo label qualities.
However, most real-world semantic segmentation datasets \cite{lin2014microsoft,cordts2016cityscapes,everingham2010pascal,zhou2019semantic} have long tail class distributions with few categories occupying the majority of pixels as illustrated in Fig.~\ref{fig:distribution}. 
And, it is well known that DCNNs trained with such long-tailed data distribution will produce predictions biased toward the dominant categories \cite{dong2018imbalanced}. This can be even more problematic for self-training, since pseudo labels are generated based on these biased model predictions. 
There exists a severe distribution mismatch between true and pseudo labels,
 especially for tail categories (see Fig.~\ref{fig:distribution}), which will harm self-training.

Recently, only very few works \cite{zou2018unsupervised,feng2020semi} attempt to address the class distribution issue in pseudo labels via sampling the same percentage of pixels for each category based on the predicted results instead of using a single confidence threshold. However, as the class distribution for the predictions has already deviated from the true distribution, the produced pseudo labels will undoubtedly still suffer from the bias.
Here, we argue that this distribution mismatch issue is a largely overlooked problem, hindering further improvements in semi-supervised semantic segmentation.

\vspace{0.1in}
\noindent
\textbf{Our Contributions.} 
In this work, we present a simple yet effective baseline method to re-distribute the biased pseudo labels, aligning their distribution with the true distribution (from the labeled set) for improving semi-supervised semantic segmentation. 

First, we highlight the distribution mismatch issue in semi-supervised semantic segmentation, formulate the task as an optimization problem, and further design a Distribution Alignment and Random Sampling (DARS) method to obtain unbiased pseudo-labeled data, matching the true class distribution. We point out that many pixels share the same confidence value (confidence overlapping) due to the over-confident issue in DCNNs~\cite{guo2017calibration}, which makes it not viable to achieve distribution matching only by thresholding. Therefore, we propose distribution alignment with class-wise thresholding and random sampling to achieve perfect distribution alignment. 

Second, during the self-training process, we contribute a progressive data augmentation and labeling strategy which gradually increases the strength of data augmentation ({\eg} the range of random scaling) and enlarges the labeling ratio. This strategy prevents the model from being overwhelmed by noisy data from an inaccurate model or strongly augmented examples at the initial stage, and avoids overfitting to high-confident pseudo-labeled easy examples through leveraging diversified augmented data and an increased number of pseudo-labeled data from an improved model.

Third, our proposed method is generic, simple and efficient, which can be seamlessly incorporated into other self-training pipelines for semi-supervised semantic segmentation by adding only a few lines of code. Albeit simple, our approach achieves surprisingly good performance compared with state-of-the-art approaches.
For Cityscapes dataset, our model gains a significant amount of performance boost of 8.89\% mIoU in the $\frac{1}{8}$ split setting, approaching the fully supervised results. 
Moreover, we also verify our method on PASCAL VOC 2012, where ours outperforms previous state-of-the-art by 4.49\% mIoU.

Finally, we further explore the performance gain in
semi-supervised semantic segmentation with the growth of unlabeled data
and find that the performance gain gradually saturates in the high-data regime. Further, we analyze the potential bottlenecks for this issue and suggest future directions, hoping to inspire more works in this direction.


\section{Related Work}
\paragraph{Supervised Semantic Segmentation.}
The introduction of fully convolutional neural networks (FCN) \cite{sermanet2013overfeat,long2015fully}
is a remarkable milestone in semantic segmentation. Most following works build upon it and either take 
advantage of multi-scale inputs \cite{chen2016attention,dai2015convolutional,farabet2012learning,lin2017refinenet,lin2016efficient,pinheiro2014recurrent},
or use feature pyramid spatial pooling \cite{liu2015parsenet,zhao2017pyramid}, or dilated convolutions
 \cite{chen2017deeplab,chen2017rethinking,chen2018encoder,li2018pyramid,wang2018understanding,yu2015multi} to improve the model, and encoder-decoder
models \cite{badrinarayanan2017segnet,chen2018encoder,li2018pyramid,ronneberger2015u}
have also been proved effective. We choose 
PSPNet \cite{zhao2017pyramid} in our main experiments for its simplicity and compelling performance, and Deeplabv2 \cite{chen2017deeplab} for a fair comparison with previous works.

\vspace{0.1cm}\noindent\textbf{Semi-Supervised Learning.}
Recently, noticeable progress has been made in the literature of semi-supervised learning,
and successful examples usually fall onto two lines of work. One is consistency training,
assuming the model’s predictions to be invariant when various perturbations are applied, 
such as UDA \cite{xie2019unsupervised}, MT \cite{tarvainen2017mean}, VAT \cite{miyato2018virtual}, 
Temporal Ensemble \cite{laine2016temporal}, Dual Student \cite{ke2019dual}. The other line of work is 
self-training \cite{xie2020self}, closely related to
entropy minimization \cite{grandvalet2005semi} and pseudo labeling \cite{berthelot2019mixmatch, sohn2020fixmatch, dong2018tri}.

In this work, we mainly focus on self-training, which often utilizes prediction confidence to assign pseudo labels to confident predictions assuming that high confidence corresponds to good accuracy. To do so, a confidence threshold is
often used to filter out low confidence unreliable predictions, and the remaining 
are constructed as pseudo-labels. Berthelot {\etal} \cite{berthelot2019mixmatch} average the predictions 
of different augmented versions of an unlabeled sample, and applies sharpening and mixup 
to generate pseudo labels. Sohn {\etal}\cite{sohn2020fixmatch} use a confidence threshold to generate pseudo-labels 
for weakly-augmented versions of unlabeled images and then train a model in the fully 
supervised way with obtained pseudo labels and stronger data augmentation. 
Xie \etal\cite{xie2020self} iteratively generate pseudo labels and train models with them. 
While these methods achieve impressive results, little attention has been paid 
to the structure and quality of pseudo-labeled data. Concurrently, \cite{berthelot2019remixmatch} and 
\cite{kim2020distribution} propose to refine pseudo labels using distribution information,
but their methods may not be well extended to pixel-level tasks like semantic 
segmentation due to the over-confident predictions and their high computational complexities for optimization. In contrast, our proposed method is simple yet efficient to handle bias in pseudo-labeling for segmentation.   


\vspace{0.1cm}\noindent\textbf{Semi-Supervised Semantic Segmentation.}
Inspired by the recent development of SSL methods in the image classification domain, 
a few works explore semi-supervised learning in semantic segmentation and show promising
results. Hung {\etal} \cite{hung2018adversarial} and Mittal {\etal} \cite{mittal2019semi} turn to 
adversarial learning, and a discriminator or a multi-label mean teacher (MLMT) branch 
is added to select reliable predictions as pseudo labels. Mendel {\etal} \cite{mendel2020semi} extend the GAN-Framework and add a secondary model as a corrector to correct the predictions 
from the segmentation model.

Consistency based methods are also frequently revisited. French {\etal} \cite{french2019semi} build upon 
\cite{zhang2018mixup} and enforce the mixed predictions and predictions of mixed inputs to be 
consistent with each other. Ouali {\etal} \cite{ouali2020semi} apply perturbations in the feature space and 
enforce consistency between predictions of different perturbation versions. 
Ke {\etal}~\cite{ke2020guided} propose a flaw detector and apply dynamic consistency constraint.

 Our proposed method is more closely related to self-training or pseudo labeling based 
methods. Concurrently, \cite{zoph2020rethinking} and \cite{chen2020semi} extend the self-training strategy
of \cite{xie2020self} from image classification to semantic segmentation. Feng {\etal} \cite{feng2020semi}
propose a class-balanced curriculum for semi-supervised semantic segmentation,
which can be viewed as the most related work to ours. 
However, these works do not 
exploit the bias in pseudo-labeling and either using a single confidence threshold 
for all classes or confining the number of samples in each class with respect to 
the biased prediction. In contrast, our method explicitly process the bias in 
pseudo-labeling and prevent it from harming self-training.




\section{Method}

\begin{figure}[t]
\begin{center}
\includegraphics[width=0.47\textwidth]{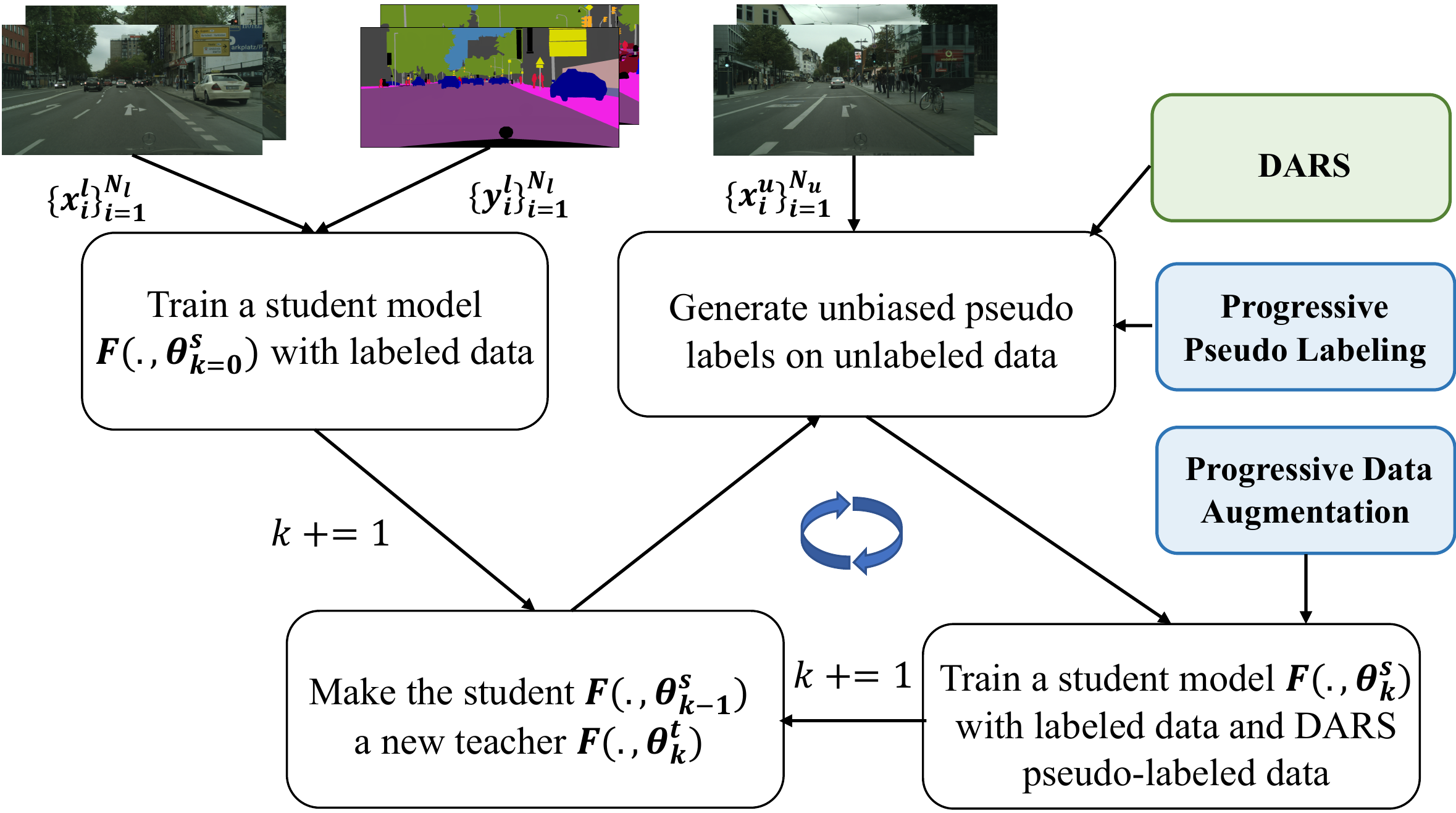}
\vspace{-0.3cm}
\end{center}
    \caption{An overview of our semi-supervised semantic segmentation framework. $k$ denotes the round index, where the round $k$=0 is the pre-training round, and rounds $k$\textgreater0 are self-training rounds.} 
\vspace{-0.4cm}
\label{fig:overview}
\end{figure}

\subsection{Overview}

In SSL, we are given a small set of labeled examples and a large set of unlabeled examples.
Let $ \mathcal{D}_l = \{(x^l_i, y^l_i)\}^{N_l}_{i=1}$ represent the $N_l$ labeled examples, 
and $ \mathcal{D}_u = \{x^u_i\}^{N_u}_{i=1}$ represent the $N_u$ unlabeled examples, where $x^l_i$ is 
the $i$-th labeled image with spatial dimensions $H \times W$, $y^l_i \in \{0,1\}^{C\times H \times W}$
is its corresponding one-hot encoded pixel-wise label map with $C$ as the number of categories, and $x^u_i$ is the $i$-th unlabeled image.

Given $ \mathcal{D}_l$ and $ \mathcal{D}_u$, our goal is to train a semantic segmentation network $F(\theta)$  with parameters $\theta$ to achieve satisfactory results on the test set with the same distribution as the training data.
An overview of our framework is shown in Fig. \ref{fig:overview}, which consists of several steps explained as follows. 

\noindent {\textbf{Step 1}}: At round $k$=0, we learn a student model $F(.,\theta^{s}_0)$ only on the $N_l$ labeled examples by minimizing
\vspace{-0.25cm}
\begin{equation}
    \vspace{-0.25cm}
   \frac{1}{N_l} \sum_{i=1}^{N_l}\mathcal{L}(y^l_i,F(x^l_i,\theta^{s}_0)),
\end{equation} 
where $ \mathcal{L}(.,.)$ denotes the cross entropy loss.

\noindent \textbf{Step 2}: At round $k$, we use the learned student model $F(.,\theta^{s}_{k-1})$ to be the teacher model $F(.,\theta^{t}_k)$, producing predictions $\{{p}^u_i\}^{N_u}_{i=1}$ for the $N_u$ unlabeled examples. Here, ${p}^u_i\in [0,1]^{C \times H \times W}$ are the network outputs after the softmax operation. Given $\{{p}^u_i\}^{N_u}_{i=1}$ and $\{y^l_i\}^{N_l}_{i=1}$, we generate the pseudo labels $\{\tilde{y}^u_i\}^{N_u}_{i=1}$ with our DARS method. 
 
\noindent \textbf{Step 3}: At round $k$, equipped with $ \widetilde{\mathcal{D}}_u = \{(x^u_i,\tilde{y}^u_i)\}^{N_u}_{i=1}$, we use both 
$ \mathcal{D}_l $, $ \widetilde{\mathcal{D}}_u$ to train a student model $F(.,\theta^{s}_k)$. The student model resumes from the teacher model and is optimized by minimizing
\vspace{-0.25cm}
\begin{equation}
\vspace{-0.2cm}
    \frac{1}{N_l} \sum_{i=1}^{N_l}\mathcal{L}(y^l_i,F(x^l_i,\theta^{s}_k)) + \frac{1}{N_u} \sum_{i=1}^{N_u}\mathcal{L}(\tilde{y}^u_i,F(x^u_i,\theta^{s}_k)).
\end{equation} 

Self-training iterates between \textbf{Step 2} and \textbf{Step 3} until no more performance gain can be achieved. During iterative training, a progressive augmentation and labeling strategy is designed to further enhance performance.

In the following, we first elaborate on our unbiased pseudo-labeled data generation with our \textbf{DARS} method in Sec. \ref{sec:phase1} . 
Then, we explain our designed strategies in data augmentation and labeling for effective training and pseudo label generation in Sec. \ref{iterative} during self-training.

\subsection{Unbiased Pseudo Label Generation} \label{sec:phase1}

To reduce noise in pseudo-labeled data and enhance their qualities, previous works either adopt a single confidence threshold~\cite{zoph2020rethinking,zhu2020improving,lian2019constructing,zhang2020transferring,sohn2020fixmatch} for all categories or a labeling ratio controlling the percentage of labeled pixels~\cite{zou2018unsupervised,feng2020semi}. 
However, both criteria suffer from the long-tail data distribution which biases pseudo-labeled data toward the dominant categories and causes a severe distribution mismatch between true and pseudo labels, thereby harming effective learning for tail categories.

In this section, we will present a very simple yet effective technique to produce unbiased high-quality pseudo-labeled data whose class distribution matches the true distribution. 
Here, we use the class distribution of the labeled data as the true distribution, since it should be representative of the real-world data under unbiased random sampling.

\vspace{0.1cm}\noindent\textbf {Problem Formulation.} Before diving into the details, we present our formulation of this problem as follows. Given the labeled data $\{(x^l_i, y^l_i)\}^{N_l}_{i=1}$ and predictions $\{{p_i^u}\}_{i=1}^{N_u}$ from the teacher model, we aim to obtain the pseudo labels $\{\tilde{y}_i^u\}_{i=1}^{N_u}$ that occupy $\alpha\%$ of all the pixels, where $\alpha$ is the labeling ratio to control the quality of pseudo-labeled pixels.  
To ensure the distribution matching and encourage pixels with high prediction confidence to have a larger possibility to be pseudo-labeled, we adopt category-specific confidence thresholds $T = \{t_i\}_{i=1}^C$ to derive the pseudo-labeled data $\{\tilde{y}_i^u\}_{i=1}^{N_u}$, where pixels with confidence scores not smaller than the corresponding class threshold are pseudo-labeled.  $T$ is derived by solving the following optimization problem,
\vspace{-0.2cm}
\begin{equation}
\begin{aligned}
    \mathop{\arg\min}_{T} \ \ & D_\text{KL}(R(\{y^l_i\}^{N_l}_{i=1}), R(\{\tilde{y}_i^u\}^{N_u}_{i=1})), \\
    \text{subject to}  \ \ &\tilde{y}_i^u = G(T,p_i^u), P(\{\tilde{y}_i^u\}_{i=1}^{N_u}) = \alpha\%.
\end{aligned}
\label{eq:ratio}
\end{equation} 
Here, given the labeled data, $R (.)$ is a frequency counting function which outputs the labeled (pseudo-labeled) pixel percentage $r_i^l$ ( or $r_i^u$) of category $i$, {\ie} $\{r_i^l\}_{i=1}^C = R(\{y^l_i\}^{N_l}_{i=1})$ or $\{r_i^u\}_{i=1}^C = R(\{\tilde{y}^u_i\}^{N_u}_{i=1})$, $D_{\text{KL}}(.,.)$ calculates the Kullback-Libeler (KL) divergence measuring the distance of two distributions. Besides, $G (.,.)$ is to generate a valid pseudo label to pixels if the confidence value is not smaller than the threshold of the corresponding category, otherwise, assign an ignore label to the pixel and $P(.)$ returns the percentage of pseudo-labeled pixels.
Notably, pixels with ignore label will not contribute to the training.

$D_{\text{KL}}$ is minimized when $R(\{y^l_i\}^{N_l}_{i=1})$ and $R(\{\tilde{y}_i^u\}^{N_u}_{i=1})$ are the same, and thus the desirable number of  pseudo-labeled pixels for each category $j$ is: 
\vspace{-0.25cm}
\begin{equation}
    \vspace{-0.25cm}
    n_j^u = \alpha \% \times N_u \times H \times W \times r_j^l. \label{eq:desired_num}
 \end{equation} 
Further, the above optimization problem is readily solvable if the confidence values are distinct with no overlapped values: $t_j$ corresponds to the $n_j^u$-th prediction value if we sort in descending order the  prediction confidence of all pixels with predicted category $j$. 

\begin{figure}
    \begin{center}
    \includegraphics[width=\linewidth]{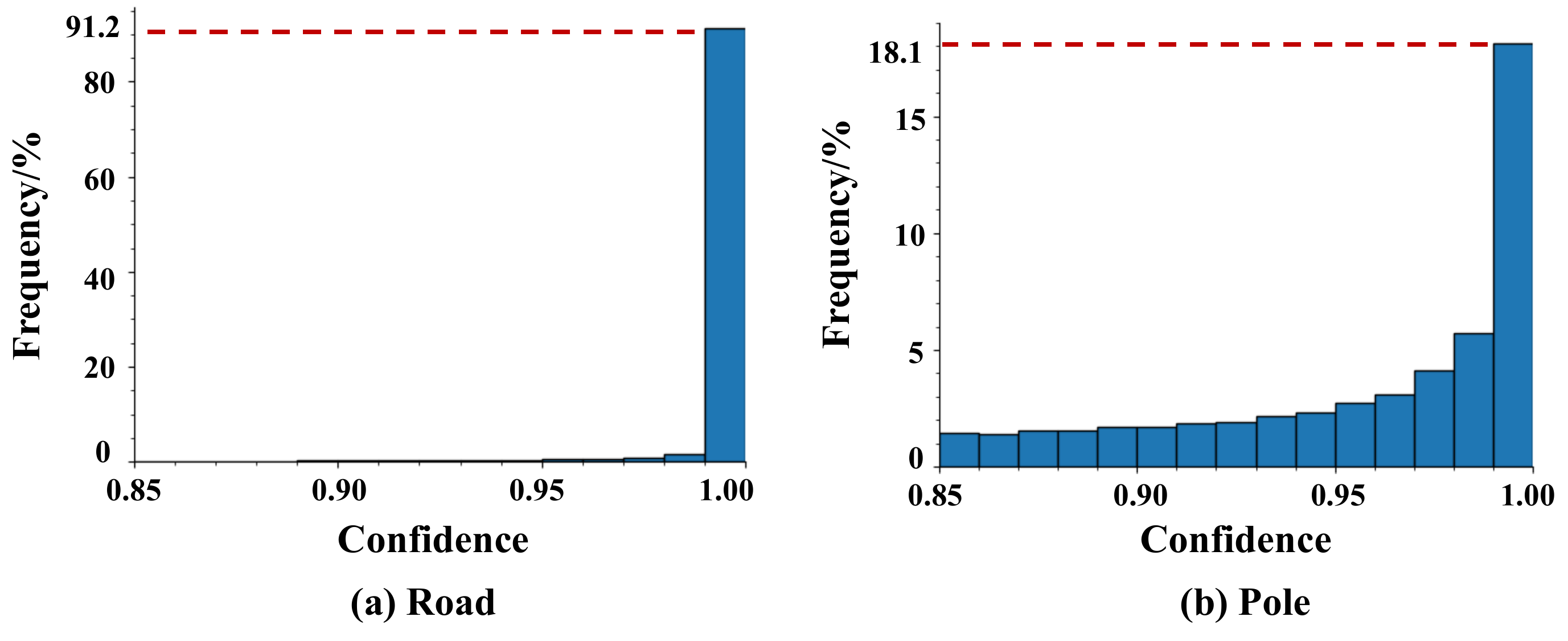}
    \end{center}
    \vspace{-0.3cm}
        \caption{Confidence overlapping issue in Cityscapes semantic segmentation task ~\cite{cordts2016cityscapes} with PSPNet50 backbone~\cite{zhao2017pyramid}. a) Histograms of confidence values for class Road; b) Histograms of confidence values for class Pole. }
    \label{fig:overlap}
    \vspace{-0.4cm}
\end{figure}

\vspace{0.1cm}\noindent\textbf{Confidence Overlapping.} However, in semantic segmentation, our observation shows many pixels have similar and indistinguishable confidence values (``confidence overlapping'' for short), which is largely due to the fact DCNNs are prone to producing over-confident prediction values~\cite{guo2017calibration}. We observe this issue is serious for head categories in semantic segmentation as shown in Fig.~\ref{fig:overlap} (a), the confidence values for the road category are distributed in a very narrow range, and the confidence of $81.3\%$ road pixels is $1$. 

This renders our previous solution to  Eq.~\eqref{eq:ratio}  not viable, which means the number of pixels after thresholding $\tilde{n}_j^u$ should be larger than $n_j^u$, and the serious confidence overlapping issue makes this not ignorable, {\ie} $\tilde{n}_j^u$ is larger than $n_j^u$ by a significant percentage. The class distribution, therefore, still deviates from the true distribution. This issue is especially severe for the head categories such as road in Cityscapes.
The overlapping in confidence values also suggests that the optimal solution to Eq.~\eqref{eq:ratio} is not unique.

Though calibration methods, among which temperature scaling \cite{hinton2015distilling,guo2017calibration,jaynes1957information} has been shown to be the most effective one for DNNs, have been studied to make DCNN's prediction calibrated and distinguishable, and there are also recent works focus on long-tailed recognition such as Focal Loss \cite{lin2017focal}, our ablation studies demonstrate that they fail to offer distribution alignment and are sub-optimal to our problem, while introducing additional cost to validate the parameters.

In the following, we present our simple yet effective method to find one solution to the above problem by alignment and sampling with a few lines of code in Algorithm~\ref{alg:distribution}.

\vspace{-0.3cm}\paragraph{Distribution Alignment and Random Sampling (DARS).} Firstly, we assume no confidence overlapping and perform distribution alignment with the optimal solution to Eq.~\eqref{eq:ratio}, as shown in Algorithm~\ref{alg:distribution} (Line~\ref{algo:line1} -- \ref{algo:line2}).
For categories that do not suffer from serious confidence overlapping, we can derive the desirable number of pixels $n_j^u$ for category $j$ by ignoring all pixels for category $j$ with confidence lower than threshold $t_j$. 
This stage resolves the distribution mismatch problem to some extent especially for tail categories which do not suffer from serious ``confidence'' overlapping issue, such as pole in Cityscapes, shown in Fig.~\ref{fig:overlap} (b).

\algnewcommand{\algorithmicforeach}{\textbf{for each}}
\algdef{SE}[FOR]{ForEach}{EndForEach}[1]
  {\algorithmicforeach\ #1\ \algorithmicdo}
  {\algorithmicend\ \algorithmicforeach}

  \renewcommand{\algorithmicrequire}{ \textbf{Input:}} 
  \renewcommand{\algorithmicensure}{ \textbf{Output:}}

\begin{algorithm} 
    \footnotesize
    \caption{DARS}
    \label{algo:two-phase re-distributing}
    \begin{algorithmic}[1]
        \Require 
        Labeled set labels $\{y^l_i\}^{N_l}_{i=1}$, network predictions on the unlabeled set $\{p_i^u\}_{i=1}^{N_u}$ and labeling ratio $\alpha$. 
        \Ensure
        Pseudo labels $\{\tilde{y}^u_i\}^{N_u}_{i=1}$  


        \State \textcolor{veg}{ \# Distribution Alignment}
        \State Calculate $\{n_i^u\}_{i=1}^C$, $\{t_j\}_{j=1}^C$ according to Eq. (\ref{eq:ratio}) and Eq. (\ref{eq:desired_num}) \label{algo:line1}
        \State Obtain initial pseudo labels $\{\tilde{y}^u_i\}_{i=1}^{N_u}$ by ignoring low confidence labels in argmax$\{p_i^u\}_{i=1}^{N_u}$ compared with $\{t_j\}_{j=1}^C$
               \label{algo:line2}
        \State \textcolor{veg}{ \# Random Sampling}
        \State Count sampling ratio: $\{s_j\}_{j=1}^C \leftarrow n^{u}_j /\tilde{n}_j^u$ \label{algo:line3}     
          \State Update $\{\tilde{y}^u_i\}_{i=1}^{N_u}$ by randomly ignoring $1-s_j$ percent pseudo-labeled pixels for each class $j$ \label{algo:line4}

    \end{algorithmic}
    \label{alg:distribution}
\end{algorithm}

\begin{figure}
    \begin{center}
    \includegraphics[width=0.8\linewidth]{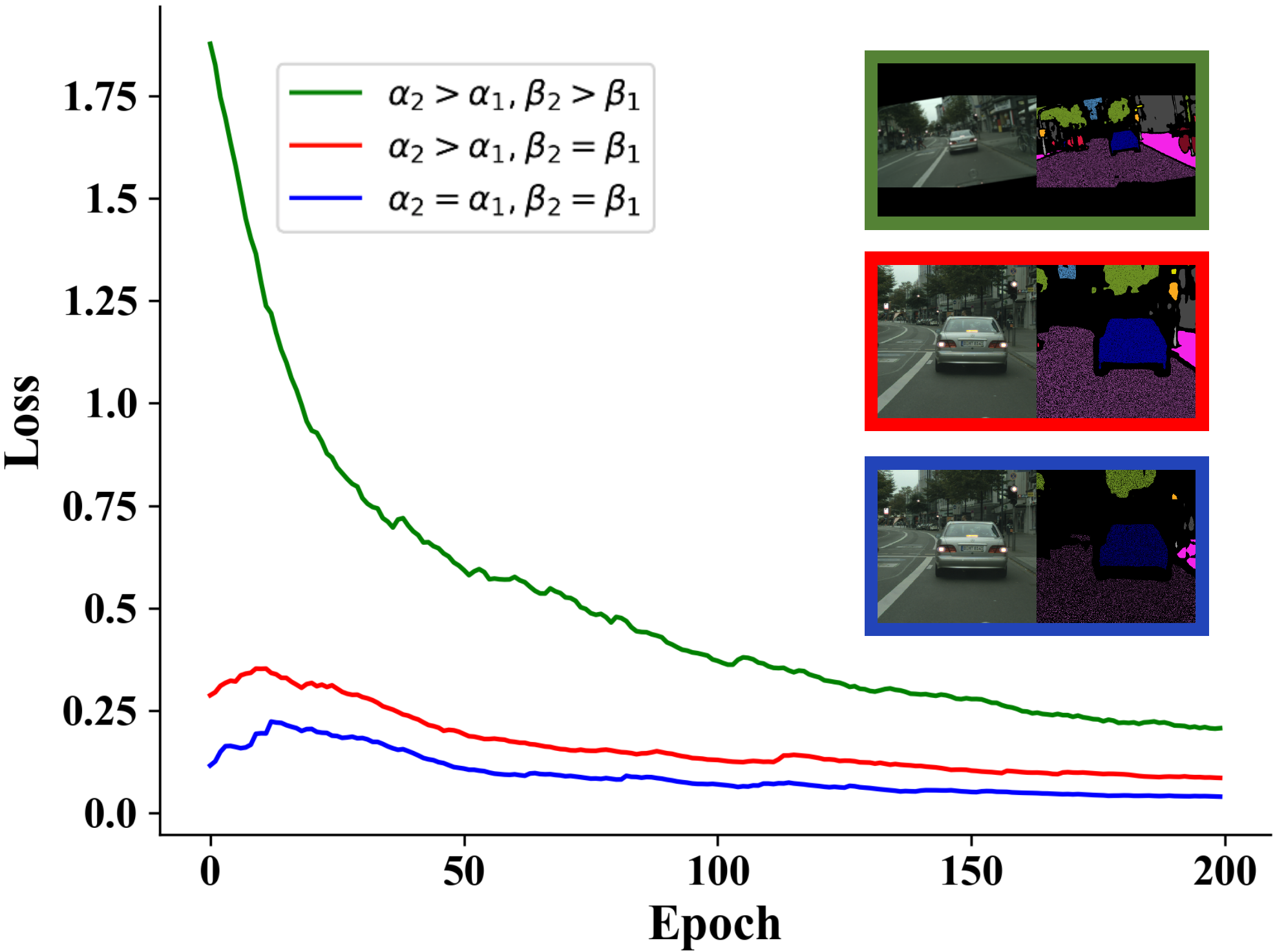}
    \end{center}
    \vspace{-0.3cm}
        \caption{Training loss of pseudo labels in iterative training ($k$=2), where $\alpha_k$ and $\beta_k$ denote the labeling ratio and the strength of data augmentation at round $k$. Right are examples of image crop and pseudo label pair for each case.}
    \vspace{-0.4cm}
    \label{fig:loss}
\end{figure}


Due to confidence overlapping, especially for head categories, the number of pixels derived after distribution alignment might be larger than the desirable number of pixels. Thus, we study how to effectively sample pixels for these categories. 
Here, we use the random sampling strategy to get the desirable number of pixels as described in Algorithm~\ref{alg:distribution} (Line~\ref{algo:line3} -- \ref{algo:line4}). The reason lies in the following folds: 1) random sampling helps redistribute the centralized high-confident pixels to different regions, effectively enlarging its spatial coverage; and 2) random sampling functions as a way of data augmentation to enhance model performance.

\vspace{-0.1cm}
\subsection{Progressive Data Augmentation and Labeling} \label{iterative}

Self-training can benefit from iterative learning. However, during the iterative learning process, if we only update pseudo labels by the latest model while keeping the labeling ratio and data augmentation magnitude the same, the training loss starts from a very low value shown in Fig.~\ref{fig:loss} (blue curve),  which implies that the model has already fit the pseudo-labeled data well, and the data cannot further improve the model performance.

Motivated by these observations, we propose to progressively enlarge the labeling ratio $\alpha$ similar to \cite{zou2018unsupervised,feng2020semi} and increase the strength of data augmentation. 
Progressively enlarging the labeling ratio helps the model harvest novel data samples without sacrificing the quality of pseudo-labeled data, benefited from an improved model. 
Though inducing novel data by enlarging the ratio could benefit model training, 
 enlarging the labeling ratio alone still provides quite little new information for iterative process since the loss curve only raises a little at the start as shown in Fig.~\ref{fig:loss} (red curve), and the model still easily fits the pseudo-labeled data which are typical high-confident easy samples.

Hence, we propose an orthogonal strategy to introduce new data samples for iterative
training through progressively increasing the magnitude of data augmentation.
Stronger data augmentations could bring unseen cases for the model and turn easy samples into challenging ones while not affecting the quality of pseudo labels, 
and thus providing new information for model updates. 

For the semantic segmentation task, random scaling is the most useful data augmentation strategy. 
Here, we also focus on strengthening the random scaling effect.
At the initial stage, we use  weak data augmentations to prevent the model from being influenced by challenging hard examples from augmentation as the model still struggles with easy examples at that time.
Then, we increase the range of scales at different self-training stages. 
Given the range of random scales $[s_\text{min}, s_\text{max}]$, the upper bound will be increased by $\beta_{max}$ and the lower bound will be decreased by $\beta_{min}$ with the new random scale range in $[(1 - \beta_{min})s_\text{min}, (1 + \beta_{max})s_\text{max}]$.  
As shown in Fig.~\ref{fig:loss} (green curve),
after increasing the magnitude, the loss for pseudo-label data starts from a relatively high point and gradually converges to zero, which suggests further model updates and improvements.

\section{Experiments}
\subsection{Experimental Setup}
\paragraph{Datasets.}
To evaluate our method, we conduct the main experiments and ablation studies on the Cityscapes dataset~\cite{cordts2016cityscapes}, which contains $5$K fine annotated images and is divided into 2975, 500 and 1525 three image sets for training, validation and test respectively.
19 urban-scene semantic classes are defined in Cityscapes for semantic segmentation. Similar to previous standards \cite{hung2018adversarial,mittal2019semi,french2019semi,olsson2020classmix,feng2020semi,mendel2020semi} in semi-supervised semantic segmentation, we randomly sample $1/8$ and $1/4$ training images to construct the labeled set, and the remaining training images consist of the unlabeled set.
To further explore the effectiveness of the proposed method, we also conduct experiments on the PASCAL VOC 2012 dataset (VOC12)~\cite{everingham2015pascal}, which provides 20 semantic classes and 1 background class. The VOC12 dataset consists of 1464 training, 1449 validation, and 1456 test images. Following previous common practice~\cite{ouali2020semi,hung2018adversarial} for semi-supervised settings, we use the official 1464 training images as labeled data and the 9k augmented set~\cite{hariharan2011semantic} as unlabeled data.

\vspace{0.1cm}\noindent\textbf{Comparison Methods.}
We denote the model trained with only the labeled set as \textbf{Baseline}, and with both the labeled set and ground truth labels of the unlabeled set as \textbf{Oracle}.

\vspace{0.1cm}\noindent\textbf{Backbone.}
It is noteworthy that most previous works employ the Deeplabv2~\cite{hung2018adversarial,mittal2019semi,french2019semi,feng2020semi,mendel2020semi} framework.
Nevertheless, we argue that exploring semi-supervised approaches based on a strong baseline could further illustrate their effectiveness and is more practical for real-world scenarios. 
Hence, we use PSPNet~\cite{zhao2017pyramid} with ResNet-50~\cite{he2016deep} as our backbone segmentation network for the main experiments.
Our reported Oracle is similar to the results reported by the paper~\cite{zhao2017pyramid}. Moreover, we also provide the results of our method with Deeplabv2 backbone for a fair comparison.

Note that we employ PSPNet instead of the top ones on leaderboards, because the SOTA methods often contain heavy engineering and parameter  tunning, accompanying large computation costs, and PSPNet is our best trade-off between reproducibility, performance and costs.

\begin{table}[htbp] 
    \centering
    \vspace{-0.2cm}
    \scalebox{0.95}{
    \begin{footnotesize}
        \setlength{\tabcolsep}{1.0mm}
        \begin{tabular}{c|c|c|c|c|c }
            \bottomrule[1pt]
            \multirow{2}{*}{Method}  & \multirow{2}{*}{Split} & \multicolumn{4}{c}{mIoU (\%)}  \\
            \cline{3-6} 
            & & Baseline & Result & Oracle & Gain \\
            \hline
            \multirow{2}{*}{Hung {\etal} \cite{hung2018adversarial}} & 1/8 & 55.5 & 58.8 & \multirow{2}{*}{67.7} & 3.3 \\
            & 1/4 & 59.9 & 62.3 &  & 2.4 \\
            \hline
            \multirow{2}{*}{Mittal {\etal} \cite{mittal2019semi}} & 1/8 & 56.2 & 59.3 & \multirow{2}{*}{65.8} & 3.1 \\
            & 1/4 & 60.2 & 61.9 &  & 1.7 \\
            \hline
            \multirow{2}{*}{CutMix \cite{french2019semi}} & 1/8 & 55.25\scriptsize{$\pm$0.66} & 60.34\scriptsize{$\pm$1.24} & \multirow{2}{*}{67.53\scriptsize{$\pm$0.35}} & 5.09 \\
            & 1/4 & 60.57\scriptsize{$\pm$1.13} & 63.87\scriptsize{$\pm$0.71} &  & 3.30 \\
            \hline
            \multirow{2}{*}{DST-CBC \cite{feng2020semi}} & 1/8 & 56.7 & 60.5 & \multirow{2}{*}{66.9} & 3.8 \\
            & 1/4 & 61.1 & 64.4 &  & 3.3 \\
            \hline
            \multirow{2}{*}{Mendel {\etal} \cite{mendel2020semi}} & 1/8 & 55.96\scriptsize{$\pm$0.86} & 60.26\scriptsize{$\pm$0.84} & \multirow{2}{*}{66.9} & 4.3 \\
            & 1/4 & 60.54\scriptsize{$\pm$0.85} & 63.77\scriptsize{$\pm$0.65} &  & 3.23 \\
            \hline
            \multirow{2}{*}{DARS (crop 361)} & 1/8 & 60.75\scriptsize{$\pm$0.35} & 69.64\scriptsize{$\pm$0.01} & \multirow{2}{*}{73.80\scriptsize{$\pm$0.34}} & \textbf{8.89} \\
            & 1/4 & 66.54\scriptsize{$\pm$0.48} & 71.30\scriptsize{$\pm$0.08} &  & 4.76 \\
            \hline
            \multirow{2}{*}{DARS (crop 713)} & 1/8 & 65.54\scriptsize{$\pm$0.34} & \textbf{72.78\scriptsize{$\pm$0.17}} & \multirow{2}{*}{76.60\scriptsize{$\pm$0.67}} & 7.24 \\
            & 1/4 & 69.22\scriptsize{$\pm$0.01}  &  \textbf{74.32\scriptsize{$\pm$0.12}} &  & \textbf{5.10} \\
            \toprule[0.8pt]
        \end{tabular}
    \end{footnotesize}
    }
    \caption{Comparison with the state-of-the-arts on Cityscapes val set. DARS uses PSPNet50 backbone.}
    \vspace{-0.5cm}

    \label{tab:sota}
\end{table}

\vspace{0.1cm}\noindent\textbf{Implementation Details.}
We implement our method using the PyTorch \cite{paszke2019pytorch} framework and set the batch size to 16.
In self-training, a batch of 16 images is composed of 8 labeled images and 8 unlabeled images, and an epoch is defined as training once on all unlabeled images as previous standard~\cite{feng2020semi}. The number of epochs in each round is 200 for Cityscapes and 50 for VOC12. During training, we employ the SGD~\cite{bottou2010large} with an initial learning rate of 0.01, momentum 0.9 and weight decay 0.0001 respectively. Also, we use a polynomial learning annealing procedure~\cite{chen2017deeplab} to schedule the learning rate. 
For data augmentation, we use random scaling, random horizontal flipping, random rotation, and random Gaussian blur. 
Due to the high computation costs of large crop size, for Cityscapes, we only take large crop size (\ie 713$\times$713) for comparison with the state-of-the-art methods, while a small crop size (\ie 361$\times$361) in our ablation studies for efficient training and evaluation. For VOC12, we take crops of 321$\times$321 as previous standard~\cite{ouali2020semi}.
To be noted, we empirically find 2 self-training rounds are enough for our implementation, and all results with iterative training (IT) undergo 2 self-training rounds. All of our results are derived by running the experiments on the same setting for \textbf{three} times. 

\vspace{-0.1cm}
\subsection{Comparison with Previous Work}
All results of DARS reported in this section contain iterative training, and are tested only by a single scale.
\vspace{-0.4cm}
\subsubsection{Cityscapes}

\vspace{-0.15cm}
\noindent\textbf{PSPNet50.}
We construct experiments to compare the proposed DARS on PSPNet50 backbone with several state-of-the-art semi-supervised semantic segmentation approaches. 
We report the result of each approach with $1/8$ and $1/4$ labeled proportions on Cityscapes. As shown in Table~\ref{tab:sota}, it is noteworthy that the performance gaps between baseline and oracle of previous works and ours are pretty close (\eg for $1/8$ split results, the gaps of all methods lie around 12\% mIoU).
Our method, though based on a stronger model, obtains more performance gains across all labeled splits. Remarkably, with only $1/8$ labeled data, our method outperforms our baseline 8.89\% in terms of mIoU and is only 4.16\% apart from the fully supervised oracle model. In addition, with a larger crop size (\ie 713), our performance is further boosted to achieve 74.32\% mIoU with $1/4$ labeled data on Cityscapes, only 2.28\% lower than the fully supervised oracle.

\noindent\textbf{Deeplabv2.}
Here, we also follow the setting of previous works \cite{hung2018adversarial,mittal2019semi,french2019semi,feng2020semi,mendel2020semi} and report the experimental results with Deeplabv2 backbone for a fair comparison. We study the setting of 1/8 split of Cityscapes, which is the most challenging setting. As shown in Table~\ref{tab:csdv2}, equipped with the same backbone, DARS still outperforms previous state-of-the-art methods by around 4\% mIoU, and is 8\% higher than the baseline and only 2.7\% to the oracle (66.9\%).

\vspace{-0.1cm}
\begin{table}[htbp] 
   \centering
   \vspace{-0.2cm}
   \begin{small}
       \begin{tabular}{c|c}
           \bottomrule[1pt]
           Method & mIoU \\
           \hline
           Deeplabv2 \cite{chen2017deeplab} & 56.2 \\
           \hline
           Hung {\etal} \cite{hung2018adversarial} & 57.1 \\
           Mittal {\etal} \cite{mittal2019semi} & 59.3 \\
           CutMix \cite{french2019semi} & 60.34 \\
           DST-CBC \cite{feng2020semi} & 60.5 \\
           Mendel {\etal} \cite{mendel2020semi} & 60.26 \\
           DARS& {\bf \color{black} 64.20} \\
           \toprule[0.8pt]
       \end{tabular}
   \end{small}
   \vspace{-0.01cm}
   \caption{Comparison with the state-of-the-arts with DeepLabv2 backbone in 1/8 split setting for Cityscapes.}
   \vspace{-0.1cm}
   \label{tab:csdv2}
\end{table}

\vspace{-0.65cm}
\subsubsection{VOC12}
\vspace{-0.2cm}
We further verify the effectiveness of our method by comparing with state-of-the-arts on VOC12 in Table~\ref{tab:voc12}. Compared with the most recent CCT method also with PSPNet50 backbone, our DARS significantly outperforms it (with 4.49\% mIoU) as well as other previous methods.

\vspace{-0.1cm}
\begin{table}[htbp] 
   \centering
   \vspace{-0.2cm}
   \begin{small}
       \begin{tabular}{ccc}
           \bottomrule[1pt]
           Method & Backbone & mIoU \\
           \hline
           GANSeg (\cite{souly2017semi}) & VGG16 & 64.10 \\
           AdvSemSeg (\cite{hung2018adversarial}) & DeepLabv2-101 & 68.40 \\
           CCT (\cite{ouali2020semi})  & PSPNet50 & 69.40 \\
           \hline
           DARS &PSPNet50 & {\bf \color{black} 73.89}  \\
           \toprule[0.8pt]
       \end{tabular}
   \end{small}
   \vspace{-0.01cm}
   \caption{Comparison with the state-of-the-arts on VOC12 val set.}
   \vspace{-0.1cm}
   \label{tab:voc12}
\end{table}

\begin{table*}[htbp!]
    \centering
    \scalebox{0.9}{
    \begin{small}
        \setlength{\tabcolsep}{0.8mm}
        \begin{tabular}{c|ccc c c c c c c c c c c c c c c c c|c|c|c}
            \bottomrule[1pt]
            Method & \rotatebox{90}{road} & \rotatebox{90}{sidewalk\ } & \rotatebox{90}{building} & \rotatebox{90}{\textcolor{blue}{wall}} & \rotatebox{90}{\textcolor{blue}{fence}} & \rotatebox{90}{\textcolor{blue}{pole}} & \rotatebox{90}{\textcolor{blue}{light}} & \rotatebox{90}{\textcolor{blue}{sign}} & \rotatebox{90}{veg} & \rotatebox{90}{\textcolor{blue}{terrain}} & \rotatebox{90}{sky} & \rotatebox{90}{\textcolor{blue}{person}} & \rotatebox{90}{\textcolor{blue}{rider}} & \rotatebox{90}{car} & \rotatebox{90}{\textcolor{blue}{truck}} & \rotatebox{90}{\textcolor{blue}{bus}} & \rotatebox{90}{\textcolor{blue}{train}} & \rotatebox{90}{\textcolor{blue}{mbike}} & \rotatebox{90}{\textcolor{blue}{bike}} & \textcolor{blue}{Tail mIoU}& mIoU & Gain \\
            \hline
            Baseline &96.7&75.4&88.2&35.2&35.0&45.8&50.3& 63.0&89.6& 53.9& 92.3& 72.1& 46.7& 90.0& 38.4& 47.9& 33.0&33.7&67.0 &47.2&60.75\scriptsize{$\pm0.35$} & 0.0 \\
            \hline
            ST & \textbf{97.4}&\textbf{78.3}& 89.5& 43.4& 38.1& 47.6& 55.7& 68.1& \textbf{90.9}& 56.6& \textbf{93.3}& \textbf{75.1}& 51.4& \textbf{91.7}& 49.0& 67.9& 38.0& 46.5& \textbf{69.9}& 54.0&65.70\scriptsize{$\pm$0.40} & 4.95 \\
            ST + TS & \textbf{97.3}& 77.7& 89.5& 43.2& 39.1& 48.3& 57.8& 68.7& \textbf{90.8}& 56.3& \textbf{93.4}& 74.9& 50.7& 91.4& 48.5& 67.9& 40.2& 46.3& \textbf{69.8}& 54.3& 65.89\scriptsize{$\pm$0.30} & 5.14\\
            CBST & 97.1& 77.6& 89.5& 42.4& 44.9& 50.2& \textbf{58.9}& \textbf{69.8}& 90.6& \textbf{57.1}& 93.1& \textbf{75.2}& \textbf{52.6}& 91.5& 49.2& 68.4& 35.5& 46.3& 69.6& 55.0& 66.29\scriptsize{$\pm$0.05} & 5.54 \\
            CBST + TS & 96.8& 77.1& 89.4& 43.7& 44.2& 50.1& 58.6& 68.9& 90.4& 55.5& 92.7& 75.0& \textbf{52.6}& 91.3& 48.8& 66.9& 42.7& \textbf{51.1}& 69.7&55.5&  66.61\scriptsize{$\pm$0.20} & 5.86 \\
            \hline
            DA + TS &97.2& \textbf{77.9}& \textbf{89.7}& \textbf{44.8}& \textbf{45.6}& \textbf{50.7}& \textbf{59.2}& 69.1& 90.6& 56.1& 93.0& \textbf{75.1}& \textbf{52.9}& 91.6& \textbf{54.8}& \textbf{69.4}& \textbf{43.1}& 48.3& 69.6& 56.7& 67.31\scriptsize{$\pm$0.12} & 6.56 \\
            DARS & 97.1 & 77.7& \textbf{89.8}& \textbf{50.2}& \textbf{46.3}& \textbf{50.8}& 58.6& \textbf{69.5}& 90.7& \textbf{57.4}& 92.8& 75.0& \textbf{52.6}& \textbf{91.8}& \textbf{57.6}& \textbf{70.3}& \textbf{44.3}&\textbf{49.9}& 69.7&\textbf{57.9}& \textbf{68.01\scriptsize{$\pm$0.12}} & \textbf{7.26} \\
            \hline
            \hline
            ST + IT & 97.5& 78.8& 89.6& 43.4& 38.5& 47.2& 55.1& 69.4& 90.9& 56.1& 93.3& 75.1& 51.4& 91.9& 49.5& 67.5& 47.3& 52.3& \textbf{70.4}& 55.2& 66.59\scriptsize{$\pm$0.14}& 5.84 \\
            CBST + IT &\textbf{97.8}& \textbf{79.2}& \textbf{90.4}& 44.6& \textbf{48.1}& \textbf{51.7}& 59.8& \textbf{70.3}& \textbf{91.1}& 58.1& \textbf{93.5}& 74.9& 52.7& \textbf{92.6}& 53.1& 70.5& 24.2& 53.6& \textbf{70.4}& 56.1& 67.20\scriptsize{$\pm$0.38}& 6.45\\
            DARS + IT & 97.2& 78.5& 90.1& \textbf{49.3}& 47.7& 50.9& \textbf{59.9}& 70.1& 90.8& \textbf{59.6}& 92.9& \textbf{75.2}& \textbf{54.4}& 92.5& \textbf{67.7}& \textbf{73.0}& \textbf{48.7}& \textbf{54.7}& 69.9& \textbf{60.6}& \textbf{69.64\scriptsize{$\pm$0.01}} & \textbf{8.89}\\
            \toprule[0.8pt]
        \end{tabular}
    \end{small}
    }
    \caption{Ablation study for different pseudo-labeling methods. The upper part reports results in a single self-training round (k=1, labeling ratio $\alpha$ =20\%), and the lower part reports results with iterative training (IT). The tail classes are highlighted in \textcolor{blue}{blue}. We make the top-2 results bold for the upper part, and top-1 bold for the lower part. Tail mIoU shows the mean IoU of tail classes. }
    \vspace{-0.2cm}
    \label{tab:component}
\end{table*}

\vspace{-0.3cm}
\subsection{Ablation Studies}
\vspace{-0.2cm}
For all of our ablation studies, we conduct experiments in the setting of 1/8 split on Cityscapes dataset, with crop size 361$\times$361 and PSPNet50 backbone.

\vspace{-0.4cm}
\subsubsection{Ablation Study for Pseudo-labeling Process}
\vspace{-0.25cm}
\noindent\textbf{Comparison of different pseudo-labeling methods.}
Since our goal is to re-distribute the biased pseudo labels towards the distribution of labeled data in iterative self-training, here, we compare our DARS method with the following different pseudo-labeling methods:
\vspace{-0.25cm}
\begin{itemize}  
    \setlength{\itemsep}{0pt} 
    \setlength{\parsep}{0pt} 
    \setlength{\parskip}{0pt}
    \item ST: the single confidence thresholding method like \cite{xie2020self,zoph2020rethinking}, regraded as the self-training baseline method;
    \item CBST: the class balanced confidence thresholding method \cite{feng2020semi,zou2018unsupervised} based on prediction results, known as the SOTA method in self-training;
    \item DA: Our proposed distribution aligning method without following random sampling to address confidence overlapping;
    \item TS: Temperature scaling, incorporating with DA, CBST or ST method to facilitate distribution alignment by calibrating model predictions as mentioned in Sec.~\ref{sec:phase1}.
\end{itemize}
\vspace{-0.25cm}

 In the upper part of Table~\ref{tab:component}, we compare our DARS with ST, ST+TS, CBST, CBST+TS, DA+TS in a single self-training round. DARS achieves 68.01\% in terms of mIoU, outperforming the single thresholding ST and class-balanced CBST by 2.31\% and 1.72\%, respectively. Equipped with temperature scaling~(TS), ST+TS, CBST + TS obtain 0.19\%, 0.32\% improvements, while our DARS is still 0.7\% superior to DA + TS strategy. This result validates TS is effective but sub-optimal in alleviating confidence overlapping and aligning pseudo label distribution.

Especially, our DARS achieves the top-2 best performance on 10 out of all 13 tail classes. Besides, we design to calculate mIoU on tail class (denoted as Tail mIoU), where our DARS further outperforms CBST 2.9\%. These experimental results demonstrate the superiority of our method on tail classes by aligning pseudo labels distribution which prevents the model from collapsing into head classes, \eg our DARS outperforms others on truck by over 8.4\% mIoU.

In the lower part of Table~\ref{tab:component}, we further compare the proposed DARS with other pseudo-labeling methods after combining iterative training. With iterative training, DARS achieves more performance gain compared with ST and CBST, and outperforms CBST+IT  by 2.44\% mIoU, leading to a significant performance boost of 8.89\% mIoU, which indicates that DARS helps resolve the bias in pseudo-labeling that may hurt iterative self-training. 


\noindent\textbf{Ablation study for distribution matching.}
While we have shown that combining Temperature scaling with previous pseudo-labeling methods lead to inferior performance compared with DARS, here, we directly compare the distribution mismatch ($D_{\text{KL}}$) between true labels from the labeled set and pseudo labels generated by different off-the-shelf techniques, including calibration methods like Temperature Scaling, Matrix Platting Scaling, Histogram Binning and approaches aim at long-tailed recognition like Focal Loss~\cite{lin2017focal}. All calibration parameters are optimized using the cross-entropy loss over the validation set as in~\cite{guo2017calibration} (T=1.27 for TS in our experiments), and hyper-parameters for focal loss adopt the advice from the paper ($\gamma$=2).

As shown in Table~\ref{tab:calib}, some techniques such as TS, Matrix Platting Scaling, and Focal Loss may help to alleviate the distribution mismatch in naive self-training (ST), but fail to eliminate it, whereas our DARS is much simpler and enables perfect distribution alignment.
\vspace{-0.2cm}
\begin{table}[htbp] 
   \centering
   \vspace{-0.18cm}
   \begin{small}
       \begin{tabular}{cc}
           \bottomrule[1pt]
           Method & $D_{\text{KL}}$ \\
           \hline
           ST &0.0491\\
           \hline
           ST + TS& 0.0357\\ 
           ST + Matrix Platting Scaling & 0.0340\\
           ST + Histogram Binning &0.4188\\
           ST + Focal Loss & 0.0396\\
           \hline
           DARS & {\bf \color{black} 0.0005}\\
           \toprule[0.8pt]
       \end{tabular}
   \end{small}
   \caption{KL divergence between the distribution of true labels and pseudo labels generated by different methods.}
   \vspace{-0.1cm}
   \label{tab:calib}
\end{table}

\vspace{-0.7cm}
\subsubsection{Ablation Study for the Progressive Strategy} 
\vspace{-0.2cm}
We explore the effectiveness of enlarging labeling ratio $\alpha$ and data augmentation magnitude (\ie $\beta_{min}$ and $\beta_{max}$) in iterative self-training, as discussed in Section~\ref{iterative}.


\begin{table}[htbp] 
    \centering
    \vspace{-0.3cm}
    \begin{small}
        \begin{tabular}{c|c|c|c|r}
            \bottomrule[1pt]
            Round & $\alpha$ (\%)  & $\beta_{min}$ & $\beta_{max}$& mIoU \\
            \hline
            1 & 20 & 0 & 0 & 68.01$\pm$0.12 \\
            1 & 20 & 0.2 & 0.5 & \textcolor{red}{(-0.47)} 67.54$\pm$0.26  \\
            \hline
            2 & 20 & 0 & 0 & \textcolor{blue}{(+0.26)} 68.27$\pm$0.12 \\
            2 & 50 & 0 & 0 & \textcolor{blue}{(+0.92)} 68.93$\pm$0.16 \\
            2 & 50 & 0.2 & 0.5 & \textcolor{blue}{(+1.63)} 69.64$\pm$0.01 \\
            \toprule[0.8pt]
        \end{tabular}
    \end{small}
    \caption{Results for different labeling ratio $\alpha$ and data augmentation setups at self-training round 0,1 at split 1/8. $\beta_{min}, \beta_{max}$ are parameters
    for data augmentation magnitude defined in~\ref{iterative}.}
    \vspace{-0.3cm}
    \label{tab:iterative}
\end{table}



As shown in Table \ref{tab:iterative}, directly using the model obtained by round $k$=1 to generate new pseudo labels and train round $k$=2 can only obtain 0.26\% mIoU improvements (\ie $\beta_{min} = \beta_{max} = 0$). However, with larger labeled ratio $\alpha = 50\%$, our round $k$=2 trained model delivers 0.92\% performance gains compared to round $k$=1 results. Furthermore, equipped our round $k$=2 self-training procedure with larger labeled ratio $\alpha = 50\%$ as well as stronger data augmentation magnitude $\beta_{min} = 0.2,\beta_{max} = 0.5$, our method obtains 1.63\% gains. These results provide strong evidence for the necessity to introduce novel yet hard examples for iterative self-training and echo our demonstrations in Section~\ref{iterative}. To validate whether we just need stronger data augmentation on each round, we show that directly applying stronger data augmentation on round $k$=1 could even cause performance degradation.

\vspace{-0.2cm}
\subsection{Additional Results and Analysis}
\vspace{-0.2cm}
\noindent\textbf{Held-out test results.} For all the reported results above, we only provide experimental results on the validation sets, following the standard in semi-supervised semantic segmentation for a fair comparison. However, to verify the generalization ability of our method and show that we do not heavily tune hyper-parameters, we also provide the results on held-out test sets.
As shown in Table~\ref{tab:held_out_test}, we obtain similar gains on held-out test sets compared with validation sets.
\vspace{-0.2cm}
\begin{table}[htbp] 
   \centering
   \vspace{-0.2cm}
   \begin{small}
       \begin{tabular}{c|r|r|r}
           \bottomrule[1pt]
           Method &  cs (crop 361) & cs (crop 713) & voc (crop 321)\\
           \hline
           baseline & 59.73 & 64.70 & 66.87  \\
           \hline
           DARS & \textcolor{blue}{(+8.10)} 67.83 & \textcolor{blue}{(+5.50)} 70.20 & \textcolor{blue}{(+4.97)} 71.84  \\
           \toprule[0.8pt]
       \end{tabular}
   \end{small}
   \caption{Held-out test set: Cityscapes (cs) 1/8 split setting; VOC12 (voc) 1.4k labeled + 9k augmented unlabeled setting.}
   \vspace{-0.1cm}
   \label{tab:held_out_test}
\end{table}

\vspace{-0.35cm}
\noindent\textbf{Improving fully-supervised models with extra unlabeled data.}
In the above experiments, we have shown the effectiveness of DARS in the low-data regime. Here, we further explore how much performance boost DARS can bring to fully-supervised models (trained with all 3K fine annotated training examples in Cityscapes, with crop size 713$\times$713).

Concretely, we utilize the given 20K coarse annotated images in Cityscapes.
However, we ignore the original coarse labels and instead generate pseudo labels for them by DARS.
To simulate the real-world scenarios, we study the impact of the number of pseudo labels, denoted as $N_p$. 
We conduct experiments with $N_p = 3K, 6K, 9K$, where we keep the training iterations to be the same.

As shown in Table~\ref{tab:coarse}, more unlabeled data increases the performance till saturation. However, the performance gain in the high-data regime is relatively small compared with the low-data regime and seems to meet a bootleneck.

\vspace{-0.15cm}
\begin{table}[htbp] 
   \centering
   \vspace{-0.2cm}
   \begin{small}
       \begin{tabular}{c|c|c|c|c}
           \bottomrule[1pt]
           $N_p$ &  0 & 3k  & 6k & 9k \\
           \hline
           mIoU & 76.60 & 78.43 & 78.64 &78.70  \\
           \toprule[0.8pt]
       \end{tabular}
   \end{small}
   \caption{Improving fully-supervised models with various number of extra unlabeled data.}
   \vspace{-0.1cm}
   \label{tab:coarse}
\end{table}

\vspace{-0.25cm}
\noindent\textbf{Analysis for potential bottlenecks in high-data regimes.}
Here, we try to analyze the bottlenecks in the high-data regime for semi-supervised semantic segmentation. Without loss of generalization, we take Cityscapes as an example to explore. 
While previous works usually group classes into head and tail classes, here we present a four-group categorization for classes in semantic segmentation from the perspective of object size and appearing frequency: (1) Classes that frequently appear (frequency$\textgreater$50\%) with large sizes (size$\textgreater$10\% of the crop area), usually the head classes, \eg road, sky; (2) Classes with high appearing frequency but small sizes, \eg pole, traffic light/sign; (3) Classes that rarely appear but with large sizes, \eg truck, bus; (4) Classes that both rarely occur and with small sizes, \eg motorcycle. 

\begin{figure}
    \begin{center}
    \includegraphics[width=1.0\linewidth]{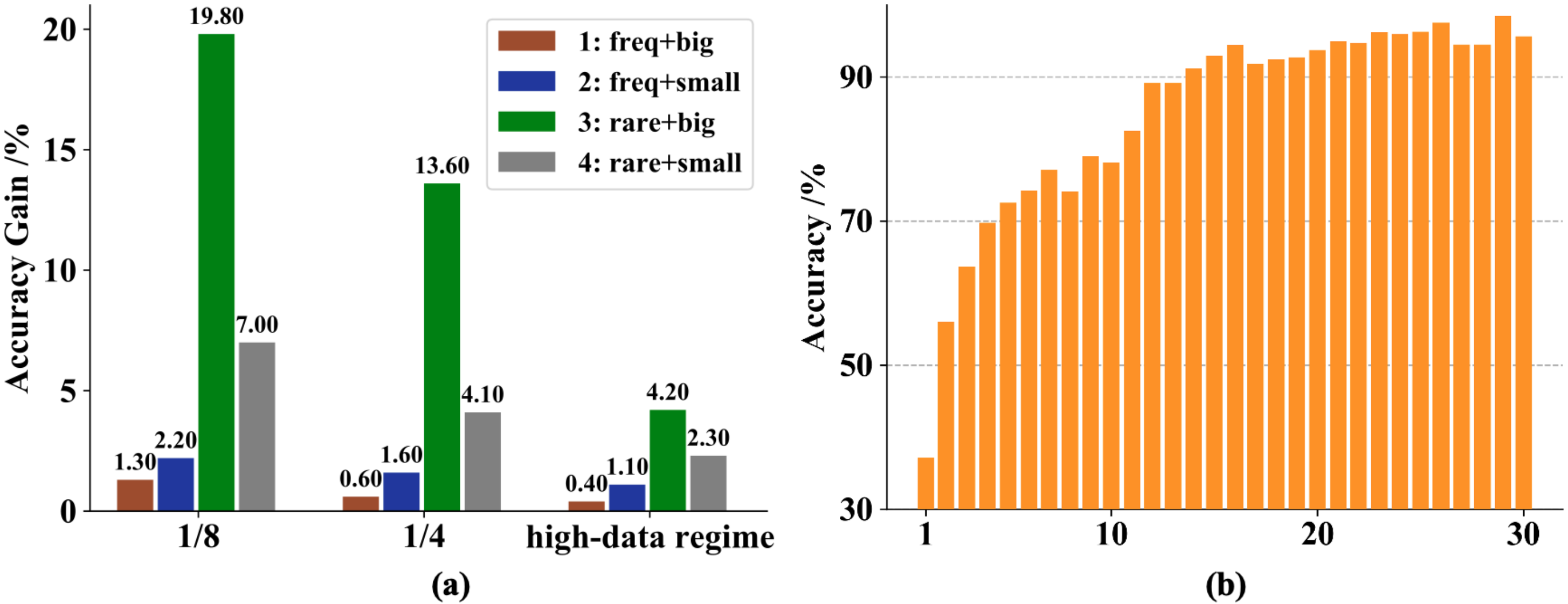}
    \end{center}
    \vspace{-0.3cm}
        \caption{(a) Performance gains of four groups in both low-data  (\ie 1/8, 1/4 split) and high-data regimes. (b) Accuracy as a function of size on Cityscapes val set. For crop size 713$\times$713=500k pixels, we use 3 different bin sizes (0.5k, 12k, 40k respectively) and each bin size for 10 bins.}
    \vspace{-0.4cm}
    \label{fig:size_acc}
\end{figure}

To better understand why only limited performance boosts could be achieved in the high-data regime, we plot the performance gain of the four groups in both low-data (\ie 1/8, 1/4 split) and high-data regimes in Fig~\ref{fig:size_acc} (a). Since group 1 classes are all head classes, it is natural that we do not harvest much boost on them. However, for the rest three groups that are all tail classes, only groups with rare occurrence attribute (\ie 3, 4) achieve noticeable gain, while group 2 achieves similar limited gains as 1. Therefore, we conclude that self-training mainly contributes to classes with rare occurrence. As a result, in the high-data regime, the occurrence of all classes increases and classes in group 3 could gradually evolve to group 1 with the growth of data, which explains the decreased performance gain in group 3 from low to high-data regime. Hence, one potential bottleneck is the gradually saturated performance for originally rare classes as training data increases. 

In order to boost performance, we would expect classes in group 2, 4 (both with small sizes) also turn into group 1, which inspires us that another bottleneck comes from the network's systematic error towards small size objects. As shown in Fig~\ref{fig:size_acc} (b), we test the baseline model's accuracy of different object size ranges. The accuracy increases as the object size grows. The network's systematic error for small objects leads to low pseudo label quality for them, and further limits self-training performance on small objects. Hence, we suggest that the second bottleneck for high-data regimes is the network's systematic error towards different object sizes. A promising direction is to develop size agnostic architecture for semantic segmentation, which we leave as future work.

\vspace{-0.2cm}
\section{Conclusion}
\vspace{-0.2cm}
We have presented a simple and yet effective DARS method to calibrate the bias in pseudo-labeling, together with a progressive data augmentation and labeling strategy for iterative self-training. Experiments on Cityscapes and VOC12 demonstrate that our simple method can outperform existing sophisticated designed approaches. We hope our formulation, encouraging results, and analysis for bottlenecks could inspire more research efforts in this direction. 

{\small
\bibliographystyle{ieee_fullname}
\bibliography{egbib}
}

\clearpage
\appendix

\definecolor{purple}{RGB}{128,64,128} 
\definecolor{pole}{RGB}{153, 153, 153}
\definecolor{trafficlight}{RGB}{250, 170, 30}
\definecolor{veg}{RGB}{107, 142, 35}
\definecolor{fence}{RGB}{190, 153, 153}
\definecolor{wall}{RGB}{102, 102, 156}
\definecolor{bus}{RGB}{0, 60, 100}
\definecolor{truck}{RGB}{0, 0, 70}
\definecolor{car}{RGB}{0, 0, 142}
\definecolor{train}{RGB}{0, 80, 100}

\centerline{\large{\textbf{Outline}}}
\vspace{0.2cm}
In this supplementary file, we first provide more results on Cityscapes in Sec.~\ref{sec:cs}: parameter analysis in Sec.~\ref{sec:param} to investigate the effects of labeling ratio and data augmentation magnitude; explanation of our consideration for data augmentation in the progressive strategy in Sec.~\ref{sec:prostra};
visualizations of pseudo labels generated by different methods in Sec.~\ref{sec:vis}; 
the qualitative results of different methods in Sec.~\ref{sec:vis2}. 
Moreover, we provide additional experimental results for the semi-supervised settings on the ScanNet dataset~\cite{dai2017scannet} in Sec.~\ref{Sec:scannet} to further explore the effectiveness of the proposed method on indoor scene segmentation. 
Finally, in Sec.~\ref{sec:uda}, we explore the importance of re-distributing biased pseudo labels by distribution aligning for unsupervised domain adaptation (UDA) on the GTA5 \cite{richter2016playing} $\rightarrow$ Cityscapes setting.

\vspace{-0.2cm}
\section{More Results on Cityscapes} \label{sec:cs}
\subsection{Parameter Analysis}\label{sec:param}
\noindent\textbf{Labeling Ratio.} Benefited from an improved teacher model, progressively enlarging labeling ratio $\alpha$ can help induce novel data while maintaining the quality of pseudo labels, and hence safely bootstrap the performance. 
Here, we present more experimental results and analysis on the Cityscapes split 1/8 at round $k$=2 to show the improvements from an enlarging labeling ratio.

\begin{table}[htbp] 
    \centering
    \vspace{-0.2cm}
    \begin{small}
        \begin{tabular}{l|c}
            \bottomrule[1pt]
            $\alpha$ (\%)  & mIoU (\%) \\
            \hline
            20 & 68.27 $\pm$ 0.12  \\
            30 & 68.54 $\pm$ 0.31  \\
            40 & 68.77 $\pm$ 0.10  \\
            50 & \textbf{68.93 $\pm$ 0.16}  \\
            60 & 68.75 $\pm$ 0.04  \\
            \toprule[0.8pt]
        \end{tabular}
    \end{small}
    \caption{Parameter analysis for labeling ratio on the Cityscapes 1/8-split at round $k$=2 with random scaling between 0.25 and 1.0.}
    \vspace{-0.3cm}
    \label{tab:labeling ratio}
\end{table}

As shown in Table \ref{tab:labeling ratio}, if we directly apply iterative training without enlarging the labeling ratio, the performance gain is quite limited (68.01\% $\rightarrow$ 68.27\%). However, as we gradually enlarge the labeling ratio, a steady performance growth is observed with the largest improvement (68.01\% $\rightarrow$ 68.93\%) achieved at $\alpha$=50\%. 

Moreover, we can observe the robustness of our progressive pseudo-labeling strategy from Table \ref{tab:labeling ratio} that noticeable performance boost could be achieved in a relatively wide range (\ie 40\% $\sim$ 60\%).

\vspace{0.1cm}\noindent\textbf{Data Augmentaion Magnitude.}
An orthogonal strategy is to progressively increase the magnitude of data augmentation. In our experiments, we focus on strengthening the random scaling factor on Cityscapes split 1/8 at round $k$=2. The range of random scaling intensity in round $k$=1 is [0.25, 1.0], which is regarded as the initial range. Afterward, we enlarge the initial range in the following self-training round, decreasing the lower bound 0.25 by $\beta_{\min}$ and increasing the upper bound 1.0 by $\beta_{\max}$. 

\begin{table}[htbp] 
    \centering
    \vspace{-0.4cm}
    \begin{small}
        \begin{tabular}{l|c|c}
            \bottomrule[1pt]
            $\beta_{\min}$ & $\beta_{\max}$ &mIoU (\%) \\
            \hline
            0.4 & 0.0 & 68.77 $\pm$ 0.12  \\
            \textbf{0.2} & 0.0 & 69.16 $\pm$ 0.03 \\
            0.0 & 0.0 & 68.93 $\pm$ 0.16  \\
            0.0 & 0.25 & 68.97 $\pm$ 0.44  \\
            0.0 & \textbf{0.5} & 69.52 $\pm$ 0.03 \\
            0.0 & 0.75 & 69.05 $\pm$ 0.06  \\
            \hline
            0.2 & 0.5 & \textbf{69.64 $\pm$ 0.01} \\
            \toprule[0.8pt]
        \end{tabular}
    \end{small}
    \caption{Parameter analysis for random scaling magnitude on the Cityscapes split 1/8 at round $k$=2, with labeling ratio=50\%.}
    \vspace{-0.3cm}
    \label{tab:scaling mag}
\end{table}

As shown in Table \ref{tab:scaling mag}, the best performance is achieved at $\beta_{min}=0.2$ and $\beta_{max}=0.5$. Notably, by enlarging the range of random scaling appropriately, a tangible performance gain is obtained (68.97\%$ \rightarrow$ 69.64\%). 

Though enlarging data augmentation magnitude to different extent leads to various performance, we observe that we can harvest performance boost in a wide range of increased data augmentation magnitude as shown in Table \ref{tab:scaling mag}, which proves the robustness of our progressive data augmentation strategy.

\begin{figure*}[t]
    \begin{center}
    \includegraphics[width=1.0\textwidth]{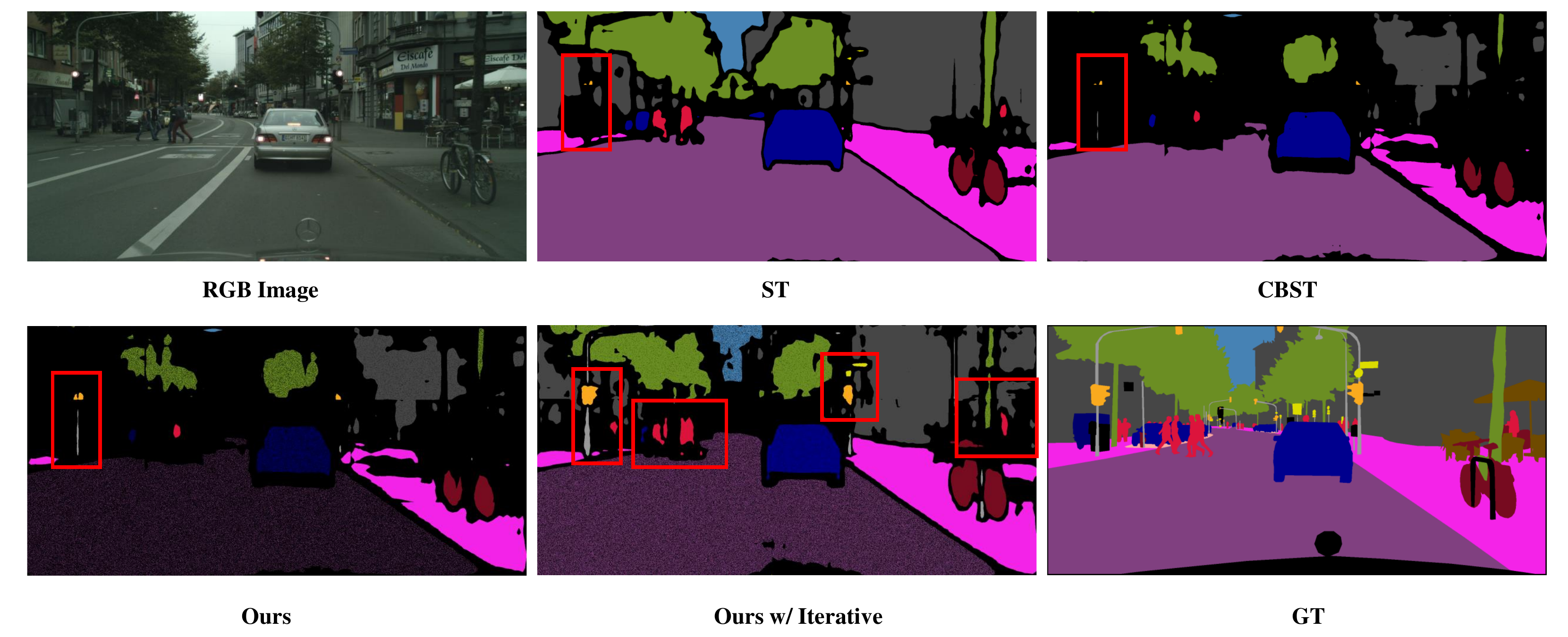}

    \end{center}
    \vspace{-0.5cm}
        \caption{Visualization of pseudo labels generated by different methods. We provide the pseudo labels of ST, CBST and ours at round $k$=1 (\ie Ours) as well as ours at round $k$=2 (\ie Ours w/ Iterative), together with the RGB image and the corresponding ground-truth. Black areas indicate the ignored region.}
    \vspace{-0.3cm}
    \label{fig:vis_pse}
    \end{figure*}
\begin{figure*}[t]
    \begin{center}
    \includegraphics[width=1.0\textwidth]{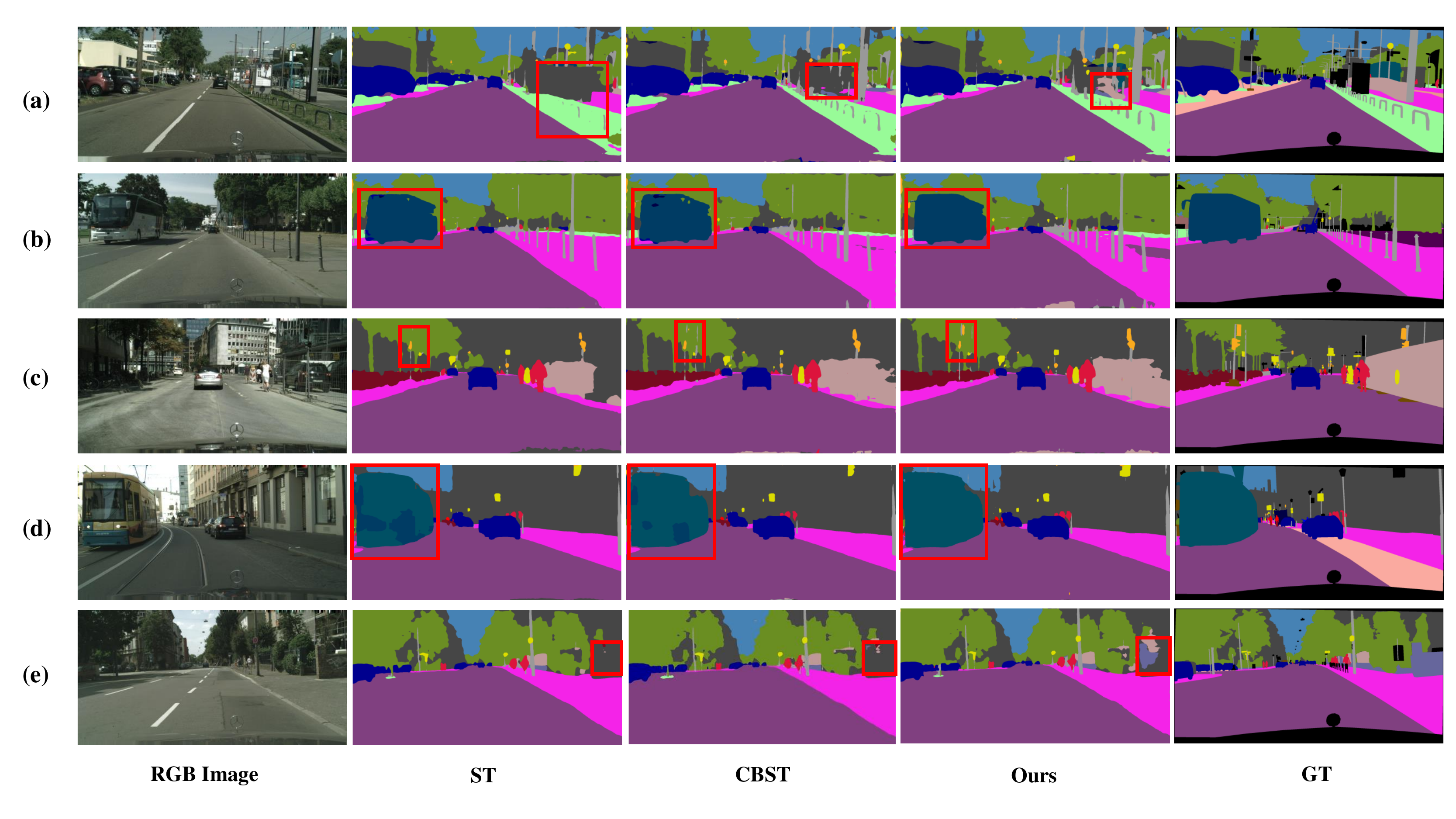}
    \vspace{-0.8cm}
    \end{center}
    \vspace{-0.3cm}
        \caption{Qualitative results of different semi-supervised methods on Cityscapes at split 1/8 at round $k$=1. Along with the RGB image and its corresponding ground-truth, we provide the results of ST, CBST and our method respectively.}
    \vspace{-0.3cm}
    \label{fig:quali}
    \end{figure*}

    \begin{table}[htbp] 
        \centering
        \begin{small}
        \vspace{-0.3cm}
            \begin{tabular}{cc}
                \bottomrule[1pt]
                Data Augmentation & mIoU \\
                \hline
                None & 70.24\\
                \hline
                Photometric Distortion& 70.84\\ 
                Random Rotation & 70.26\\
                Random Scaling & 74.36\\
                \toprule[0.8pt]
            \end{tabular}
        \end{small}
        \caption{Effectiveness of different data augmentation methods in semantic segmentation.}
        \vspace{-0.3cm}
        \label{tab:calib}
     \end{table}

\begin{table*}[htbp]
    \centering
    \scalebox{0.95}{
    \begin{small}
        \setlength{\tabcolsep}{0.8mm}
        \begin{tabular}{c|ccc c c c c c c c c c c c c c c c c c|c|c}
            \bottomrule[1pt]
            Method & \rotatebox{90}{wall} & \rotatebox{90}{floor} & \rotatebox{90}{cabinet} & \rotatebox{90}{bed} & \rotatebox{90}{chair} & \rotatebox{90}{\textcolor{blue}{sofa}} & \rotatebox{90}{table} & \rotatebox{90}{door} & \rotatebox{90}{\textcolor{blue}{window}} & \rotatebox{90}{\textcolor{blue}{bookshelf}} & \rotatebox{90}{\textcolor{blue}{picture}} & \rotatebox{90}{\textcolor{blue}{counter}} & \rotatebox{90}{desk} & \rotatebox{90}{\textcolor{blue}{curtain}} & \rotatebox{90}{\textcolor{blue}{refriger}} & \rotatebox{90}{\textcolor{blue}{shower curt}} & \rotatebox{90}{\textcolor{blue}{toilet}} & \rotatebox{90}{\textcolor{blue}{sink}} & \rotatebox{90}{\textcolor{blue}{bathtub}} & \rotatebox{90}{\textcolor{blue}{others}} &  mIoU & Gain \\
            \hline
            Baseline &75.1& 81.3& 47.0& 63.5& 56.7& \textbf{43.8}& 59.3& 48.6& 44.4& 55.6& 28.3& 34.5& 45.4& 44.8& 47.3& 53.4& 76.6& 52.5& 62.8& 36.1&52.8&0.0 \\
            \hline
            ST & 76.9& 82.3& 48.5& 66.4& 57.1& 42.3& 60.0& 51.5& 46.1& 54.3& 33.6& 36.1& 47.9& 48.0& 53.7& 55.6& 79.3& 53.6& 69.2& 36.9& 55.0& 2.2 \\
            CBST &77.2& \textbf{83.5}& 49.1& 66.7& \textbf{57.7}& 41.9& 62.5& \textbf{52.5}& 46.1& \textbf{57.1}& 34.2& \textbf{37.0}& 50.4& 46.4& 54.3& 50.7& 80.3& 54.8& 68.9& 37.2& 55.4& 2.6 \\
            \hline
            DARS (Ours) & \textbf{77.5}& 83.4&\textbf{50.2}& \textbf{67.0}& 57.5& \textbf{43.8}& \textbf{63.0}& 52.3& \textbf{46.6}& 56.7& \textbf{35.2}& 35.3& \textbf{50.6}& \textbf{49.4}& \textbf{54.7}& \textbf{58.3}& \textbf{80.6}& \textbf{54.9}& \textbf{69.3}& \textbf{38.4}&  \textbf{56.2}& \textbf{3.4}\\
            \toprule[0.8pt]
        \end{tabular}
    \end{small}
    }
    \caption{Comparisons of different semi-supervised approaches on the $1/8$ split of the ScanNet dataset at round $k$=1 with labeling ratio $\alpha$=50\%. The tail classes are highlighted in \textcolor{blue}{blue}. We make the best performance result bold for each class. Single scale test is adopted for all methods here.}
    \vspace{-0.2cm}
    \label{tab:compare}
\end{table*}

\begin{table}[htbp] 
    \centering
    \scalebox{1}{
    \begin{small}
        \setlength{\tabcolsep}{1.0mm}
        \begin{tabular}{c|c|c|c|c|c }
            \bottomrule[1pt]
            \multirow{2}{*}{Method}  & \multirow{2}{*}{Split} & \multicolumn{4}{c}{mIoU (\%)}  \\
            \cline{3-6} 
            & & Baseline & Result & Oracle & Gain \\
            \hline
            \multirow{2}{*}{DARS (Ours) } & 1/8 & 52.84 & 56.58 & \multirow{2}{*}{61.69} & 3.74 \\
            & 1/4 & 56.61  & 58.35  &  & 1.74 \\
            \toprule[0.8pt]
        \end{tabular}
    \end{small}
    }
    \caption{Our results on ScanNet with iterative training on split 1/8 and 1/4.}
    \vspace{-0.5cm}
    \label{tab:ours scannet}
\end{table}

\subsection{Data augmentation in the progressive strategy} \label{sec:prostra}
Here, we explain why only random scaling is considered in the progressive strategy. Our previous empirical experiments showed that random scaling is the most useful data augmentation method for semantic segmentation. To be specific, we have conducted experiments on the Cityscapes dataset with different data augmentation methods. Specifically, we trained a PSPNet50 using all 2975 fine-annotated training images with a crop size of 361$\times$361 (half-resolution training). 
For data augmentation methods, we consider photometric distortion (brightness, contrast, saturation, and hue), random rotation, and random scaling following common setups in previous work [56]. We employed one of the three data augmentation methods or none of them, respectively, and report the performance on the validation set. As shown in Table RB1, random scaling could bring significant performance boosts, whereas photometric distortion or random rotation could only bring limited gains. We will add this analysis to the supplementary material upon publication.

Moreover, applying too strong magnitudes for data augmentation methods like brightness and rotation might influence data distribution.
Hence, we only consider random scaling in the progressive strategy. However, we believe other data augmentation methods like mixup could also be incorporated into our progressive strategy to further boost performance and we hope our idea of progressively increasing data augmentation magnitude for iterative training could benefit future research.

\subsection{Visualization of Pseudo Labels} \label{sec:vis}
To provide more information about our approach, we visualize the pseudo labels generated by our method as well as conducting comparisons with methods such as ST and CBST on the Cityscapes dataset. 
    
As shown in Fig. \ref{fig:vis_pse}, pseudo labels form ST and CBST are often overwhelmed by the majority classes like \textbf{\textcolor{purple}{road}} and \textbf{\textcolor{veg}{vegetation}}. And the tail class objects are often ignored in their pseudo labels, such as the \textbf{\textcolor{pole}{pole}} and \textbf{\textcolor{trafficlight}{traffic light}} in the red box. As a result, the label distribution of their pseudo labels is extremely biased towards the dominant classes.

In contrast, with our distribution alignment and random sampling strategy to deal with the confidence overlapping phenomenon, the percentage of dominant classes are reduced and the pseudo labels are re-distributed to cover a large spatial area.
Besides, 
our method successfully pseudo-labels the tail classes such as the \textbf{\textcolor{pole}{pole}} and \textbf{\textcolor{trafficlight}{traffic light}} in the red box at round $k$=1 (see Ours in Figure~\ref{fig:vis_pse}).

Further, 
when we enlarge the labeling ratio to 50\% at round $k$=2, the quality of our pseudo labeled data is further enhanced. More tail class objects are pseudo-labeled and incorporated into our pseudo labels, as shown by the red boxes of Ours (w/ Iterative) in Figure~\ref{fig:vis_pse}.

\subsection{Qualitative Results}\label{sec:vis2}
In this section, we provide qualitative results of the semi-supervised semantic segmentation methods on the Cityscapes dataset. Concretely, we compare our results with ST and CBST methods at round $k$=1.

As shown in Fig. \ref{fig:quali}, previous methods mainly have two failure modes in segmenting tail classes: (1) 
they tend to leave out some tail classes like \textbf{\textcolor{fence}{fence}}, \textbf{\textcolor{trafficlight}{traffic light}} and \textbf{\textcolor{wall}{wall}} ({\eg} in the red box areas, the \textbf{\textcolor{fence}{fence}} is missing in (a), one \textbf{\textcolor{trafficlight}{traffic light}} is lost in (c), and the \textbf{\textcolor{wall}{wall}} is completely unrecognized in (e));
(2) they suffer from the confusion with similar classes and mistake tail class object as other classes. For instance, in (b), part of the \textbf{\textcolor{bus}{bus}} is mistaken as \textbf{\textcolor{veg}{vegetation}}, \textbf{\textcolor{truck}{truck}} or \textbf{\textcolor{car}{car}}, and in (d), some part of the \textbf{\textcolor{train}{train}} is misclassified as \textbf{\textcolor{bus}{bus}}.

Thanks to our distribution alignment and sampling strategy to calibrate the bias, our method can alleviate the above two issues and thus outperforms ST and CBST on tail classes significantly. 
As shown in Fig. \ref{fig:quali}, our method can successfully segment the tail class objects as in (c) and recognize most tail class areas ({\eg} the \textbf{\textcolor{fence}{fence}} in (a) and the \textbf{\textcolor{wall}{wall}} area in (e)). Moreover, our method significantly improves the model's ability to handle the confusion between similar classes and give consistent and correct predictions as in (b) and (d).

\section{Additional Experiments on ScanNet} \label{Sec:scannet}
To further demonstrate the transferability and broad applicability of our method, we evaluate it on the indoor scene dataset, ScanNet~\cite{dai2017scannet}. To be noted, we do not tune the hyper-parameters on the ScanNet dataset to show the generality of our method.

\vspace{-0.3cm}
\paragraph{Dataset.} ScanNet is an RGB-D  dataset collected from 1,513 indoor scenes. 
{For the 2D semantic segmentation task, ScanNet contains 19,466 RGB images for training and 5,436 images for validation with a resolution of 1296$\times$968. In our semi-supervised setting,  1/8 ({\ie} 1/8-split) and 1/4 ({\ie} 1/4-split) of the images are randomly chosen from the training set to serve as the labeled set.
}
Pixel-level annotations for the following 21 object classes are provided: \textsl{wall, floor, cabinet, bed, chair, sofa, table, door, window, bookshelf, picture, counter, desk, curtain, refrigerator, shower curtain, toilet, sink, bathtub, other furniture, and void (the ignore category)}. 

\vspace{-0.3cm}
\paragraph{Implementation Details} 
We follow the same experimental setup as the Cityscapes dataset, except that the number of epochs is set to 20 for each training round and a crop size of 481$\times$481 is adopted. Also, since the variance in our experiments is rather small as shown in this supplementary file, we only run one experiment for each setting on ScanNet to save the computational cost.


\begin{table*}[htbp!] 
    \centering
    \begin{footnotesize}
        \setlength{\tabcolsep}{0.8mm}
        \begin{tabular}{c|c|ccc c c c c c c c c c c c c c c c c | c} 
            \bottomrule[1pt]
            Method & Backbone & \rotatebox{90}{road} & \rotatebox{90}{sidewalk\ } & \rotatebox{90}{building} & \rotatebox{90}{\textcolor{blue}{wall}} & \rotatebox{90}{\textcolor{blue}{fence}} & \rotatebox{90}{\textcolor{blue}{pole}} & \rotatebox{90}{\textcolor{blue}{light}} & \rotatebox{90}{\textcolor{blue}{sign}} & \rotatebox{90}{veg} & \rotatebox{90}{\textcolor{blue}{terrain}} & \rotatebox{90}{sky} & \rotatebox{90}{\textcolor{blue}{person}} & \rotatebox{90}{\textcolor{blue}{rider}} & \rotatebox{90}{car} & \rotatebox{90}{\textcolor{blue}{truck}} & \rotatebox{90}{\textcolor{blue}{bus}} & \rotatebox{90}{\textcolor{blue}{train}} & \rotatebox{90}{\textcolor{blue}{mbike}} & \rotatebox{90}{\textcolor{blue}{bike}} &mIoU \\
            \hline
            CyCADA \cite{hoffman2018cycada} &  VGG-16& 85.2 &37.2& 76.5& 21.8& 15.0& 23.8 &22.9& 21.5& 80.5& 31.3& 60.7& 50.5& 9.0 &76.9& 17.1& 28.2& 4.5 &9.8 &0.0 &35.4 \\
            \hline
            ASN \cite{tsai2018learning} & ResNet-101 &86.5                 & 25.9                       & 79.8                     & 22.1                 & 20.0                  & 23.6                 & 33.1                                    & 21.8                 & 81.8                & 25.9                    & 75.9                & 57.3                                     & 26.2                                    & 76.3                & 29.8                                    & 32.1                                  & 7.2                            & 29.5                                    & 32.5                                   & 41.4  \\
            CLAN \cite{luo2019taking}   &ResNet-101 & 87.0                 & 27.1                       & 79.6                     & 27.3                 & 23.3                  & 28.3                 & 35.5                           & 24.2        & 83.6                & 27.4                    & 74.2                & 58.6                                     & 28.0                                    & 76.2                & 33.1                                    & 36.7                                  & 6.7                                     & 31.9                                    & 31.4                                   & 43.2          \\
            ADVENT \cite{vu2019advent}  &ResNet-101 &89.4 &33.1& 81.0 &26.6 &26.8 &27.2 &33.5 &24.7 &83.9 &36.7 &78.8 &58.7 &30.5& 84.8& 38.5 &44.5 &1.7& 31.6 &32.4 &45.5 \\
            CAG-UDA \cite{zhang2019category} & ResNet-101& 90.4 &\textbf{51.6}& 83.8& 34.2 &27.8 &38.4& 25.3& \textbf{48.4}& 85.4& 38.2& 78.1 &58.6& \textbf{34.6}& 84.7& 21.9& 42.7& \textbf{41.1}& 29.3& 37.2 &50.2 \\
            RPT \cite{zhang2020transferring} &ResNet-101 & 89.7& 44.8& \textbf{86.4}& \textbf{44.2}& \textbf{30.6}& 41.4 &\textbf{51.7}& 33.0& \textbf{87.8}& \textbf{39.4}& \textbf{86.3} &65.6 &24.5& \textbf{89.0} &36.2 &46.8& 17.6& \textbf{39.1}& \textbf{58.3}& 53.2 \\
            \hline
            CBST \cite{zou2018unsupervised} & WideResNet-38 & 89.6 &\textbf{58.9} &78.5& 33.0& 22.3 &41.4 &48.2 &39.2& 83.6 &24.3& 65.4& 49.3& 20.2& 83.3 &\textbf{39.0} &\textbf{48.6}& 12.5& 20.3 &35.3 &47.0 \\
            PyCDA \cite{lian2019constructing} & WideResNet-38 & \textbf{92.3} &49.2& 84.4& 33.4 &30.2 &33.3 &37.1& 35.2& 86.5& 36.9& 77.3 &63.3& 30.5& 86.6 &34.5& 40.7 &7.9& 17.6& 35.5& 48.0 \\
            CRST \cite{zou2019confidence} & WideResNet-38 & \textbf{91.7} &45.1& 80.9& 29.0 &23.4 &\textbf{43.8} &47.1& 40.9& 84.0& 20.0& 60.6& 64.0 &31.9& 85.8& \textbf{39.5}& \textbf{48.7} &\textbf{25.0}& 38.0& 47.0 &49.8 \\
            \hline
            CyCADA$^*$ \cite{hoffman2018cycada} & ResNet-50  & 85.6& 37.6& 81.7& 34.3& 20.2& 35.8& 41.4& 31.7& 85.2& 37.8& 74.0& \textbf{66.5}& 24.5& 83.5& 24.6& 19.0& 0.0& 30.1& 26.7&  44.2\\
            DARS (TD) &ResNet-50& 90.6& 50.8& \textbf{88.0}& \textbf{43.4}& \textbf{33.9}& \textbf{48.3}& \textbf{53.4}& \textbf{50.2}& \textbf{87.0}& \textbf{46.2}& \textbf{80.9}& \textbf{71.6}& \textbf{34.4}&\textbf{87.3}& 33.1& 40.6& 6.6& \textbf{45.6}& \textbf{53.4}& \textbf{55.0} \\
            \toprule[0.8pt]
        \end{tabular}
    \end{footnotesize}
    \caption{Adaptation results from GTA5 $\rightarrow$ Cityscapes. The tail classes are highlighted in \textcolor{blue}{blue}. We make the top-2 performance results bold for each class. CyCADA$^*$: we re-implements CyCADA on our PSPNet-50 framework. }
    \vspace{-0.1cm}

    \label{tab:uda sota}
\end{table*}

\begin{table}[htbp]
    \vspace{-0.2cm}
    \centering
    \begin{small}
        \begin{tabular}{l|c|c|c}
            \bottomrule[1pt]
            Method  & $D_{\text{KL}}$ & mIoU (\%) & Gain (\%) \\
            \hline
            Baseline & / & 43.43 $\pm$ 0.01 & 0.0 \\
            \hline
            ST & 0.0727 &48.04 $\pm$ 0.69 & +4.61 \\
            CBST & 0.0196 & 48.74 $\pm$ 0.16 & +5.31 \\
            DARS (SD) & 0.1558 &47.01 $\pm$ 0.42 & +3.58 \\
            DARS (TD) & 0.0006& \textbf{51.01 $\pm$ 0.05} & \textbf{+7.58} \\
            \toprule[0.8pt]
        \end{tabular}
    \end{small}
    \caption{GTA5 $\rightarrow$ Cityscapes results at round $k$=1. $D_{\text{KL}}$ indicates the KL divergence between the distribution of pseudo labels and Cityscapes real labels.}
    \vspace{-0.5cm}
    \label{tab:uda}
\end{table}

\vspace{-0.1cm}
\vspace{0.1in}
\noindent
\textbf{Main Results.}
We compare the proposed DARS method with the single thresholding method \cite{zoph2020rethinking,zhu2020improving,lian2019constructing,zhang2020transferring,sohn2020fixmatch} (ST) and the class balance thresholding method \cite{zou2018unsupervised,feng2020semi} (CBST) considering on the 1/8-split setting at round $k$=1 without iterative training.   
As shown in Table \ref{tab:compare}, the proposed simple DARS method achieves 56.2\% mIoU on the validation set, surpassing ST and CBST method, which reiterates the superiority of our proposed method.
To be noted, our method introduces little computational cost in comparison with the compared approaches. 

Further, we report the final results of the proposed method with iterative training at split 1/8 and 1/4 in Table \ref{tab:ours scannet}. Notably, our method achieves 58.35\% in terms of mIoU with only 1/4 labeled data, which is very close to the fully-supervised results of 61.69\%.

\vspace{-0.2cm}
\paragraph{Analysis} We notice that the performance gain achieved by self-training is relatively small on the ScanNet dataset in comparison with the Cityscapes dataset. 
We mainly attribute this to the difference between indoor and urban scenes. While urban scenes usually have similar structures ({\eg} road is always at the bottom and the sky at the top), indoor scenes tend to have large variance and complex spatial relationships which impose obstacles for pseudo-labeling that relies on models trained with only a small set of labeled data. Exploring the 3D structure for semi-supervised learning in indoor scene parsing have the potential to address these difficulties which will be our future work. We believe our method could also be incorporated into other methods to further boost the performance for semi-supervised in-door scene parsing.
Also, we barely finetune the hyper-parameters like labeling ratio and data augmentation magnitude to save time and computational costs since our main purpose for experiments on ScanNet is to show the broad applicability of our method with superiority to previous self-training methods.

\section{Unsupervised Domain Adaptation Setting} \label{sec:uda}
In this section, we further conduct experiments on the more challenging unsupervised domain adaptation setting, in order to confirm our major insight about the importance of semantic-level distribution alignment in pseudo-labeling. 

While we do not have the labeled set for the target domain to obtain the true label distribution, for comparison fairness with other methods, we could not perform DARS for generating unbiased pseudo labels. Instead, we use this setting to study the relationship between the extent of distribution mismatch in pseudo labels (\ie KL divergence with target label distribution) and the performance boost.

We compare the following pseudo-labeling methods: 
\begin{itemize} 
    \setlength{\itemsep}{0pt}
    \setlength{\parsep}{0pt}
    \setlength{\parskip}{0pt}
    \item ST: the single confidence thresholding method like \cite{xie2020self,zoph2020rethinking}, regraded as the self-training baseline method;
    \item CBST: the class balanced confidence thresholding method \cite{feng2020semi,zou2018unsupervised}, which actually uses confidence to estimate the target label distribution;
    \item DARS (SD): DARS using the source label distribution as the target label distribution;
    \item DARS (TD): DARS using the target label distribution counted on the validation set of the target domain. 
\end{itemize}

\vspace{-0.5cm}
\paragraph{Dataset.}
We follow \cite{hung2018adversarial,vu2019advent,yang2020adversarial} to consider the popular synthetic-to-real adaptation task: GTA5 $\rightarrow$ Cityscapes. The GTA5 dataset~\cite{richter2016playing} provides 24,966 images with pixel-wise labels. We use the 19 classes of GTA5 in common with the Cityscapes for adaptation. 
Moreover, we take advantage of image translation and use images translated by CyCADA \cite{hoffman2018cycada} in GTA5 for training. 

\vspace{-0.3cm}
\paragraph{Implementation Details} 
We also follow the same experimental setup as the Cityscapes dataset, except that the number of epochs is set to 10 for the pre-training round on GTA5 and a crop size of 713$\times$713 is adopted.

\vspace{0.1in}
\noindent
\textbf{Main Results.}
As shown in Table \ref{tab:uda}, the smaller the KL divergence between the distribution of pseudo labels and target labels is, the better performance is achieved, which highly validates our motivation to re-distribute biased pseudo labels.
Moreover, it is noteworthy that when the pseudo labels achieve perfect distribution alignment with true distribution (\eg DARS (TD)), it could achieve much more performance gain than other pseudo-labeling (\eg 4\% mIoU higher than DARS (SD), 2.27\% higher than CBST in a single round). 

Further, we report the results with iterative training of DARS (TD) in comparison with previous works in Table \ref{tab:uda sota}. We claim that the comparison is not fair since DARS (TD) utilizes the target label distribution from the validation set, and we show the encouraging and superior performance (\ie 55.0\% mIoU) of it only to highlight the importance of distribution aligning in pseudo-labeling for unsupervised domain adaptation settings, hoping to inspire more works in this direction.


\end{document}